\documentclass[conference]{IEEEtran}
\usepackage{times}

\usepackage{xcolor}
\definecolor{prettypurple}{RGB}{116, 1, 113}

\usepackage[
colorlinks=true,
linkcolor=prettypurple,
urlcolor=prettypurple,
citecolor=prettypurple,
bookmarks=true,
]{hyperref}
\usepackage[numbers,sort&compress]{natbib}
\usepackage{multicol}
\usepackage[dvipsnames]{xcolor}
\usepackage{algorithm}
\usepackage{algpseudocode}
\usepackage{blindtext}
\usepackage{makecell}
\usepackage{mdframed}
\usepackage{booktabs}
\usepackage{amsmath}
\usepackage{amssymb}
\usepackage{graphicx}
\usepackage{caption}
\usepackage{xspace}
\usepackage{booktabs,tabularx}
\usepackage{multirow} 

\usepackage{amsmath}
\usepackage{amssymb}
\usepackage{amsfonts}
\usepackage{graphicx}
\usepackage{wrapfig}
\usepackage{caption}
\usepackage[most]{tcolorbox}
\usepackage{booktabs}
\let\labelindent\relax
\usepackage{enumitem}
\usepackage{threeparttable}
\usepackage{caption}
\usepackage{subcaption}
\usepackage{listings}

\newenvironment{tightlist}%
{\begin{list}{$\bullet$}{%
    \setlength{\topsep}{0in}
    \setlength{\partopsep}{0in}
    \setlength{\itemsep}{0in}
    \setlength{\parsep}{0in}
    \setlength{\leftmargin}{1.5em}
    \setlength{\rightmargin}{0in}
}
}%
{\end{list}
}
\DeclareMathOperator*{\argmin}{argmin}

\usepackage{titlesec}
\usepackage[commandnameprefix=ifneeded]{changes}
\definechangesauthor[color=black]{rishabh}

\makeatletter
\let\IEEEorig@ltocsection\l@section
\newcommand{\apptocoff}{\let\l@section\@gobbletwo}
\newcommand{\apptocon}{\let\l@section\IEEEorig@ltocsection}
\makeatother

\renewcommand{\theparagraph}{\arabic{paragraph}}

\titleformat{\paragraph}[runin]{\bfseries}{\theparagraph.\ }{0em}{}[.]
\titlespacing{\paragraph}{0pt}{.2ex plus 1ex minus .2ex}{1em}

\begin{document}

\title{\textbf{TACTIC}: \underline{\textbf{T}}actile \underline{\textbf{a}}nd Vision \underline{\textbf{C}}onditioned Con\underline{\textbf{t}}act Centr\underline{\textbf{i}}c \underline{\textbf{C}}ontrol for Whole-Arm Manipulation}

\author{
\IEEEauthorblockN{
Rishabh Madan\textsuperscript{1},
Angchen Xie\textsuperscript{2},
Samantha Saak\textsuperscript{1},
Andres Blanco\textsuperscript{1},
Dohyeok Lee\textsuperscript{1},\\
Sarah Grace Brown\textsuperscript{1},
Yunting Yan\textsuperscript{1},
Mark Zolotas\textsuperscript{3},
Jose Barreiros\textsuperscript{3},
Tapomayukh Bhattacharjee\textsuperscript{1}
}

\IEEEauthorblockA{
\textsuperscript{1}Cornell University 
\textsuperscript{2}Carnegie Mellon University 
\textsuperscript{3}Toyota Research Institute
}
}

\twocolumn[{%
\renewcommand\twocolumn[1][]{#1}%

\maketitle
\thispagestyle{empty}
\pagestyle{empty}

\vspace{-10pt}
\begin{center}
    \captionsetup{type=figure}
    \includegraphics[width=\textwidth]{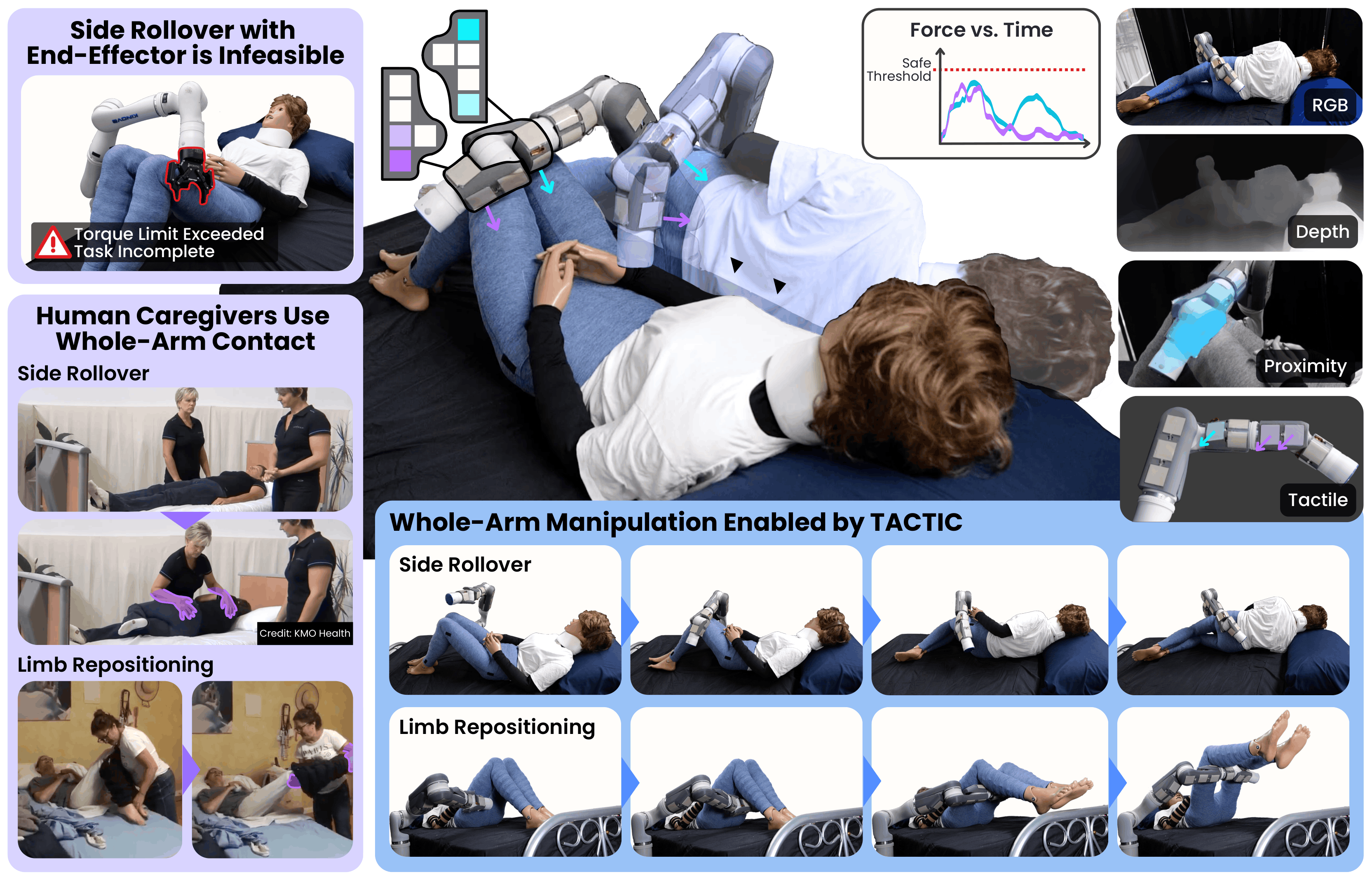}
    \captionof{figure} {\textbf{TACTIC} enables robots to perform contact-rich manipulation by utilizing the entire arm surface. While end-effector-centric control is often infeasible and unsafe under high payloads, TACTIC fuses multimodal observations such as RGB-D, proximity masks, and distributed tactile sensing into a contact-centric representation to enable whole-arm manipulation.
    We demonstrate its effectiveness on manipulation tasks, including side rollover and limb repositioning on a life-size manikin.}
    \label{fig:front}
\end{center}
\vspace{-2mm}
}]

\begin{abstract} Whole-arm manipulation involves direct contact with the environment while the robot completes a task by distributing contact across multiple links as contacts form, slide, and break. This setting breaks common implicit assumptions in many learning-based manipulation pipelines: arm configuration tightly couples motion and contact forces, contact state is partially observed under occlusion, and purely learned rollouts can become physically inconsistent under distribution shift because many multi-link contact configurations are sparsely represented in the data. To address this, we propose TACTIC (Tactile and Vision Conditioned Contact-Centric Control), a receding-horizon controller for whole-arm manipulation. TACTIC uses a contact-centric hybrid predictive model that combines RGB-D, distributed tactile sensing, and a compact 2D proximity representation. The model couples a learned, action-conditioned latent dynamics model with analytical kinematics through contact Jacobians, enabling rollouts of future contact configurations and interaction forces. TACTIC integrates these rollouts into a sampling-based MPC planner with contact-aware action sampling: contact Jacobian-based projections steer sampled action sequences toward force-modulating directions, and objectives defined over predicted proximity and interaction forces trade task progress against whole-arm force regulation. We evaluate TACTIC in simulation against state-of-the-art model-based and model-free methods, and perform ablations that isolate the contribution of each design choice. Across experiments, TACTIC consistently outperforms other methods. We further demonstrate real-world performance on a robot with distributed tactile sensing across three whole-arm manipulation tasks that require multi-contact trajectories: turning over and repositioning a manikin, and goal-reaching in a 3D dynamic maze. Website: \url{emprise.cs.cornell.edu/tactic}.
\end{abstract}

\IEEEpeerreviewmaketitle

\newcommand{\R}{\mathbb{R}}
\newcommand{\norm}[1]{\left\lVert #1 \right\rVert}

\newcommand{\vct}[1]{\mathbf{#1}}
\newcommand{\mtx}[1]{\mathbf{#1}}
\newcommand{\set}[1]{\mathcal{#1}}

\newcommand{\obs}{o}
\newcommand{\obsGoal}{o_g}
\newcommand{\act}{\vct{u}}
\newcommand{\actSet}{\set{U}}
\newcommand{\horizon}{H}
\newcommand{\ndof}{n}

\newcommand{\rgb}{I}
\newcommand{\depth}{D}
\newcommand{\forces}{\vct{f}}
\newcommand{\nsens}{N_s}

\newcommand{\q}{\vct{q}}
\newcommand{\qd}{\dot{\vct{q}}}
\newcommand{\dt}{\Delta t}

\newcommand{\Fsafe}{F_{\mathrm{safe}}}

\newcommand{\Cost}{\mathcal{J}}

\newcommand{\cc}{\mathrm{cc}}
\newcommand{\obscc}{o^{\cc}}
\newcommand{\proxmask}{M}

\newcommand{\lat}{\vct{z}}
\newcommand{\latHat}{\hat{\vct{z}}}
\newcommand{\enc}{\phi}
\newcommand{\decForce}{\psi_{\mathrm{contact}}}
\newcommand{\forcesHat}{\hat{\vct{f}}}

\newcommand{\Nroll}{N}
\newcommand{\nomAct}{\boldsymbol{\mu}}
\newcommand{\covAct}{\boldsymbol{\Sigma}}
\newcommand{\pert}{\boldsymbol{\delta}}
\newcommand{\accNoise}{\boldsymbol{\eta}}

\newcommand{\histlen}{L}
\newcommand{\contactw}{\rho}

\newcommand{\cset}{\mathcal{K}}
\newcommand{\ncon}{K}
\newcommand{\nrm}{\vct{n}}
\newcommand{\Jn}{\mtx{J}^{n}}
\newcommand{\fmag}{f}
\newcommand{\Fsat}{F_{\mathrm{sat}}}
\newcommand{\ridge}{\varepsilon}

\newcommand{\Pforce}{\mtx{G}^{\mathrm{force}}}
\newcommand{\Pnull}{\mtx{G}^{\mathrm{null}}}
\newcommand{\shapeMat}{\mtx{S}}

\newcommand{\Fhyb}{\mtx{F}}
\newcommand{\Fr}{\mtx{F}_{k}}
\newcommand{\Fl}{\mtx{F}_l}

\providecommand{\wVal}{\lambda_V}

\providecommand{\ckin}{c_{\mathrm{kin}}}
\providecommand{\cforce}{c_{\mathrm{force}}}
\providecommand{\cgoal}{c_{\mathrm{goal}}}
\providecommand{\Vfn}{V_\theta}

\newcommand{\latGoal}{\vct{z}_g}

\newcommand{\SymbolGlossary}{%
\begin{table}[t]
\centering
\scriptsize
\setlength{\tabcolsep}{2pt}
\renewcommand{\arraystretch}{0.92}
\begin{tabularx}{\columnwidth}{@{}
  >{\centering\arraybackslash}p{0.18\columnwidth} >{\raggedright\arraybackslash}X
  @{\hspace{3pt}}
  >{\centering\arraybackslash}p{0.18\columnwidth} >{\raggedright\arraybackslash}X
@{}}
\toprule
\textbf{Symbol} & \textbf{Meaning} & \textbf{Symbol} & \textbf{Meaning} \\
\midrule
$\ndof$                   & arm DoF                                  & $\horizon$                                      & MPC horizon \\
$\q_t,\;\qd_t$            & joint pos./vel.                          & $\Nroll$                                        & MPPI rollouts \\
$\act_t \in \R^{\ndof}$   & cmd joint vel. (input)                   & $\nomAct_{t+i}$                                 & MPPI mean at $t+i$ \\
$\actSet$                 & admissible actions (e.g., limits)        & $\covAct$                                       & MPPI covariance \\
$\obs_t$                  & multimodal observation                   &
\shortstack[c]{$\Cost(\act;\,\obs_{1:t},$\\$\obsGoal)$} & MPC objective \\
$\obsGoal$                & goal observation                         & $\forces_t \in \R^{\nsens}$                     & taxel forces ($\nsens$ sensors) \\
$\obscc_t$                & contact-centric obs. from $\obs_t$       &
\shortstack[c]{$\decForce(\latHat_{t+i})$}        & decoded binary contacts \\
$\proxmask_t$             & proximity mask                           & $\Fsafe$                                        & safe-force limit \\
$\lat_t,\;\latHat_{t+i}$  & latent / predicted latent                & $\cset_t$ ($|\cset_t|=\ncon_t$)                 & active contacts \\
$\enc(\cdot)$             & encoder                                  & $\Jn_t$                                         & contact Jacobian (normal) \\
$\Fl$                     & latent dynamics model                    & $\fmag_{k,t}$                                   & force magnitude (contact $k$, time $t$) \\
$\Fr$                     & analytical kinematics model              & $\Fsat$                                         &  saturation force\\
\bottomrule
\end{tabularx}
\caption{Glossary of notation used in this paper.}
\end{table}%
}

\addtocontents{toc}{\protect\setcounter{tocdepth}{-2}}
\addtocontents{toc}{\protect\apptocoff}
\vspace{-1mm}
\section{Introduction}
Recent progress in robot learning spanning vision-language-action (VLA) models~\cite{zitkovich2023rt}, diffusion-based policies~\cite{barreiros2025careful}, and world models~\cite{assran2025v} has improved performance across a wide range of manipulation tasks~\cite{vuong2023open}. By leveraging large datasets, these methods acquire priors over scenes and skills and often yield desired behavior with minimal task-specific engineering. However, a gap remains: these approaches struggle in \emph{contact-rich manipulation} where success requires reasoning about and regulating interaction forces. This gap is especially pronounced in \emph{whole-arm manipulation}, where the robot deliberately uses multiple arm links to make, break, and regulate contact while executing a task.

Whole-arm contact is fundamental to human manipulation: people use the forearm and upper arm to brace, hold, guide, and reposition objects and bodies, distributing forces over a larger area to improve stability and comfort~\cite{grice2013whole}. Robots require similar capabilities in contact-rich settings such as assistive caregiving and physical human–robot interaction (pHRI). These challenges also arise when manipulating large, unwieldy, or deformable objects in cluttered environments, where contacts form and redistribute along the arm. Whole-arm manipulation introduces two key challenges. First, trajectory generation is hard because arm configuration tightly couples motion and forces: it determines which links make contact with the environment, where contacts form and break, and how forces evolve over time. Second, data is scarce for the regimes that matter most for whole-arm manipulation: diverse multi-contact configurations and force-regulating behaviors are difficult to collect and remain sparsely represented in existing datasets. \looseness=-1

Existing approaches often address only a subset of these challenges, typically emphasizing either learned models or analytical structure. Purely data-driven models can be brittle because widely used datasets are dominated by end-effector-centric interactions and lack
the diversity of multi-link contact configurations needed to generalize to whole-arm behaviors. Even with
tactile supervision, learned predictors can produce physically inconsistent contact and force estimates
(e.g., unrealistic force magnitudes) under distribution shift. Conversely, approaches that rely solely
on analytical models~\cite{jain2013reaching, 7029969} provide explicit mechanisms for safety but often
fail to capture the multimodal, partially observed interactions that arise in contact-rich whole-arm
manipulation. Relying on either paradigm alone can therefore yield systems that either capture rich
interactions but struggle to regulate safety-related objectives under distribution shift, or enforce
safety without generating complex whole-arm behaviors.

Reliable whole-arm manipulation requires reasoning about simultaneous contacts distributed along the arm. Because contact locations and interaction forces are difficult to infer from joint torques alone, we use distributed tactile sensing to observe contact directly. Yet sensing contact is not enough; the controller must represent, predict, and act on contact information to balance task progress with whole-arm force regulation. We use receding-horizon sampling-based MPC as our control framework, for two reasons: receding-horizon control enables continual replanning as contacts form, break, and shift across the arm, while sampling-based optimization can evaluate diverse candidate actions under nonconvex and contact-dependent dynamics. Within this framework, we ground our approach in three design principles.  First, the state should be contact-centric and multimodal: we fuse RGB-D with distributed
tactile sensing and a compact 2D proximity representation that summarizes where the arm is near or in
contact with the scene (including a person) and how this region evolves over time. Second, planning
should use contact-aware action sampling: in constrained multi-contact settings, standard sampling
schemes (e.g., MPPI, CEM) are inefficient at discovering force-modulating actions. We use tactile
sensor-derived contact locations to compute contact Jacobians, which allow us to bias sampled action
sequences toward directions that can regulate interaction forces. Third, prediction and evaluation
should be hybrid and contact-aware: we couple a learned, action-conditioned latent dynamics model with
analytical kinematics via contact Jacobians to roll out contact-relevant dynamics and interaction forces,
and we evaluate candidate trajectories with objectives defined over predicted proximity and forces to
trade task progress against whole-arm force regulation.

Built on these principles, we propose \textbf{TACTIC} (\textit{Tactile and Vision Conditioned Contact-Centric Control}), a receding-horizon controller for whole-arm manipulation. TACTIC uses a \emph{contact-centric} state representation by coupling RGB-D, distributed tactile sensing, and a proximity map around the arm. To explore efficiently under multi-contact constraints, TACTIC uses
sampling-based MPC with \emph{contact-aware action sampling}, biasing candidate action sequences toward force-modulating directions via contact Jacobian projections. For prediction and trajectory scoring, TACTIC couples an action-conditioned latent dynamics model (trained on the contact-centric state) with analytical kinematics through contact Jacobians to predict the evolution of proximity and interaction forces, and optimizes objectives defined over these predictions. We evaluate TACTIC through simulation
studies that compare against competitive baselines and include ablations to quantify the impact of each design choice. We further demonstrate real-world performance on a robot with distributed tactile sensing in two contact-rich whole-arm manipulation tasks involving a life-size manikin, side rollover and limb repositioning, and in a 3D dynamic maze goal-reaching task that requires complex multi-contact trajectories. Finally, we develop and release open hardware for a mirrored Kinova Gen3-7DoF \cite{kinova_gen3_page} teleoperation interface and a WiFi-based distributed piezoresistive tactile sensing suite for the Gen3 arm.

\noindent The main contributions of this paper are as follows:
\begin{itemize}
    \item[\textbf{C1.}] \emph{Contact-centric} state representation for whole-arm manipulation that fuses RGB-D, distributed tactile sensing, and a compact 2D proximity map.
    \item[\textbf{C2.}] \emph{Contact-aware action sampling} for sampling-based MPC that uses contact Jacobian projections to bias exploration toward force-modulating directions in constrained multi-contact settings.
    \item[\textbf{C3.}] \emph{Hybrid predictive model} that couples learned, action-conditioned latent dynamics with analytical kinematics via contact Jacobians to predict contact dynamics and forces, and scores trajectories with objectives defined over predicted proximity and interaction forces.
    \item[\textbf{C4.}] Simulation and real-robot experiments with ablations, demonstrating improved performance relative to purely learned and purely analytical baselines across multi-contact tasks.
    \item[\textbf{C5.}] Open hardware for a Kinova Gen3 exoskeleton and a WiFi-based distributed piezoresistive tactile sensing suite, including design files, assembly instructions, and code.
\end{itemize}

\vspace{-1mm}
\section{Related Work}

\textbf{Whole-arm manipulation.}
Recent research in whole-arm manipulation (WAM) leverages distributed tactile skins to localize and regulate contact \cite{xu2024cushsense, si2023robotsweater, goncalves2022punyo, liu2022warm}. These capabilities have been applied to manipulating heavy or bulky objects \cite{gliesche2021geometry, barreiros2025learning}, navigating clutter \cite{jain2013reaching}, and providing physical \cite{madan2025prioritouch} and social \cite{block2021six} assistance. Prior studies show that allowing whole-arm contact improves reachability in cluttered environments \cite{jain2013reaching}. Furthermore, example-guided reinforcement learning can acquire robust whole-body manipulation strategies for large objects on compliant humanoid hardware with tactile sensing \cite{barreiros2025learning}. Despite these advances, many WAM systems are not designed for whole-arm pHRI, where the robot must infer task-relevant contact state from multimodal observations (vision, proprioception, and tactile) under occlusion and regulate interaction forces online. In contrast, our approach leverages multimodal sensing to adapt contact-rich whole-arm behaviors and incorporates explicit force-related objectives within MPC to support safer physical interaction.

\textbf{Visuotactile representations for manipulation.}
Contact-rich manipulation often demands feedback beyond vision due to occlusion, compliance, and rapidly changing contact modes. Visuotactile representations, therefore, fuse vision and touch to infer task-relevant contact state and improve robustness during interaction (e.g., self-supervised multimodal representations for contact-rich tasks and visuotactile policies that adapt after contact)~\cite{lee2019making,calandra2018more,chen2022visuo}. These representations have also been paired with model-based control to close the loop on contact, including tactile servoing and predictive control frameworks that use learned dynamics from touch (and, in some cases, vision) to plan under contact uncertainty~\cite{tian2019manipulation,xu2024letac}. In parallel, model-based learning for contact-rich manipulation learns task-specific contact models or action-conditioned world models and embeds them into planning to improve predictive accuracy under contact~\cite{okada2024contact,zhu2025irasim,khorrambakht2025worldplanner,assran2025v}. However, many such pipelines remain primarily vision-centric and do not explicitly exploit distributed tactile supervision or analytical kinematic structure, which can be valuable for grounding contact reasoning and enforcing force-aware objectives in pHRI. In contrast, our approach uses a contact-centric visuotactile representation and integrates it with analytical kinematics via joint state propagation and contact Jacobians for contact-aware prediction and planning. \looseness=-1

\vspace{-1mm}

\setlength{\abovedisplayskip}{6pt}
\setlength{\belowdisplayskip}{6pt}

\begin{figure*}
    \centering
    \includegraphics[width=\textwidth]{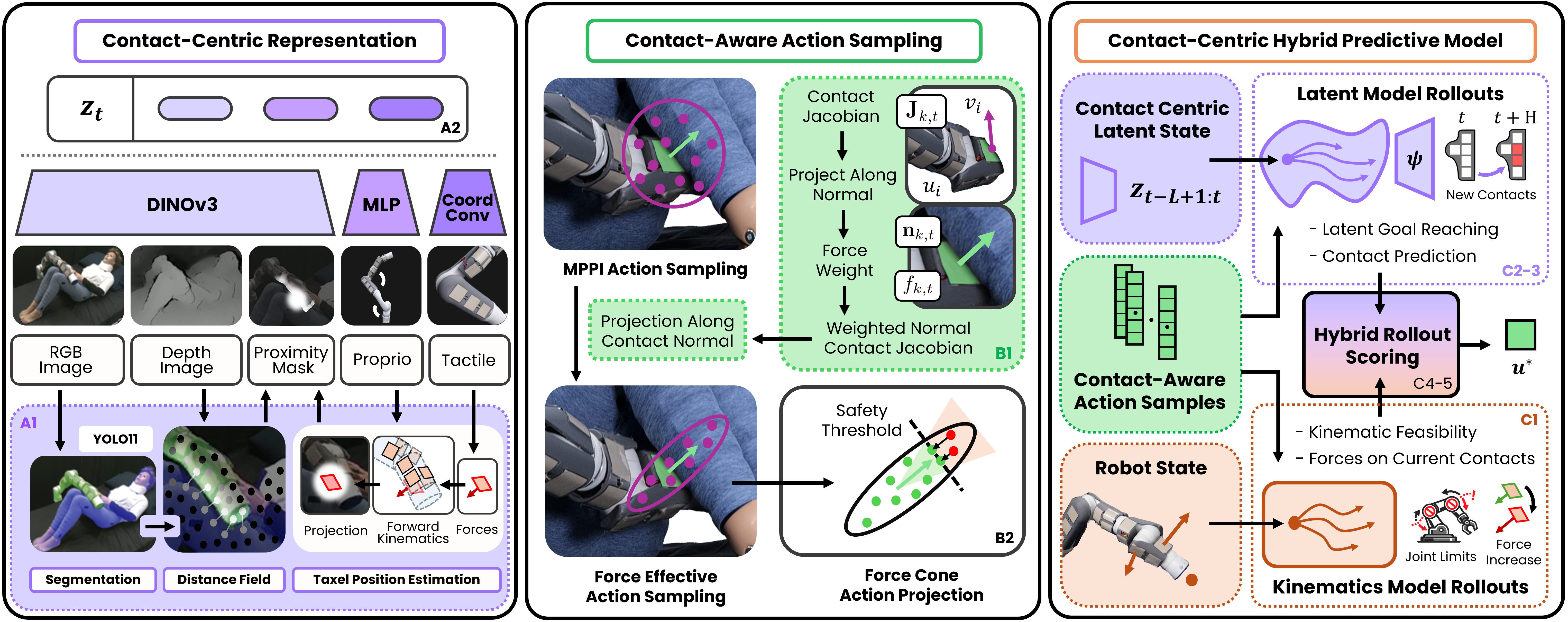}
    \caption{Overview of TACTIC. \textbf{A. Contact-Centric Representation} encodes multimodal history into a latent state $z_t$. \textbf{B. Contact-Aware Action Sampling} utilizes tactile-derived contact Jacobians ($\mathbf{J}_{k,t}$) to bias MPPI exploration toward force-modulating directions. \textbf{C. Contact-Centric Hybrid Predictive Model} evaluates candidate sequences by combining an analytical kinematics branch for physical consistency with a learned latent dynamics model for predicting contact and force evolution.}
    \label{fig:placeholder}
    \vspace{-5mm}
\end{figure*}

\section{Problem Formulation}
\label{sec:problem_formulation}

\SymbolGlossary

We consider whole-arm manipulation where the robot accomplishes
a task by moving its arm while making, breaking, and regulating contact along its links. The robot is an $n$-DoF manipulator with joint configuration $\q_t\in\mathbb{R}^n$, joint
velocity $\dot{\mathbf{q}}_t\in\mathbb{R}^n$, and control input given by commanded joint velocity
$\act_t\in\mathbb{R}^n$.
The robot receives a multimodal observation
\(
o_t \;=\; (I_t,\; D_t,\; \forces_t,\; \q_t,\; \dot{\mathbf{q}}_t),
\label{eq:obs}
\)
where $I_t$ represents an RGB image, $D_t$ represents an aligned depth map, and $\forces_t\in\mathbb{R}^{N_s}$ are distributed tactile readings
over $N_s$ taxels (\underline{ta}ctile pi\underline{xels}) along the arm. We are given a goal observation $o_g$
and safety constraints, including maximum allowable contact force bounded by a threshold $\Fsafe$. Our objective is to reach the goal while respecting these safety limits.

We use \emph{sampling-based} receding-horizon MPC, and instantiate the online optimizer with Model-Predictive Path Integral (MPPI) \cite{williams2017model}. At each time step $t$, MPPI approximates the solution of a finite-horizon constrained problem by sampling candidate action sequences, evaluating
their costs, and updating the nominal sequence toward lower-cost rollouts. We then execute only the
first action and replan at the next step with updated observations:
\begin{equation}
\mathbf{u}^\star_{t:t+H-1}
= \argmin_{\mathbf{u}_{t:t+H-1}\in\mathcal{U}^H}
\;\mathcal{J}\!\left(\mathbf{u}_{t:t+H-1};\, o_{1:t},\, o_g\right)
\label{eq:receding_horizon_problem}
\end{equation}

\section{TACTIC for Whole-arm Manipulation}
This section outlines the core design decisions of TACTIC as a sampling-based controller for whole-arm manipulation.

\vspace{-2mm}
\subsection{Building a Contact-Centric Representation}
\label{sec:contact_centric_repr}

TACTIC learns an action-conditioned dynamics model from vision, distributed force data, and proprioception. From the raw observation $\obs_t$ in Eq.~\ref{eq:receding_horizon_problem}, we compute a proximity mask $\proxmask_t$ and construct the augmented observation
\(
\obscc_t \;=\; ( \rgb_t,\; \depth_t,\; \proxmask_t,\; \forces_t,\; \q_t,\; \qd_t ).
\)

\paragraph{Proximity mask $\proxmask_t$}
Tactile readings are spatially sparse, available only upon contact, and localize contact on the arm rather than in the scene. Yet many tasks require dense, anticipatory reasoning about \emph{where in the scene} contact is occuring or likely to occur next. We define the proximity mask $\proxmask_t$ as an image-aligned 2D representation of robot-object proximity that grounds impending and active contact in the scene frame, highlighting arm surface regions near or contacting the object. First, we segment the RGB image into robot ($\Omega_r$) and object ($\Omega_o$) masks using finetuned YOLOv11~\cite{khanam2024yolov11}. Using camera intrinsics, we unproject pixels in $\Omega_r$ and $\Omega_o$ into 3D point sets $P_r$ and $P_h$, respectively, then compute each robot point's distance to the closest object point using kd-tree nearest-neighbor queries~\cite{maneewongvatana1999analysis}. We map these distances to pixel intensity (near $\rightarrow$ bright, far $\rightarrow$ dark) to obtain a dense visual proximity field. Finally, we fuse tactile data by rendering a binary taxel indicator image $\proxmask^{\mathrm{tax}}_t$, initialized to zero. For each taxel with force exceeding $f_{\min}$, we project its 3D pose via forward kinematics onto the image and render a unit-valued blob of radius $r$ at the projected location. The final $\proxmask_t$ is the per-pixel maximum of the visual proximity field and $\proxmask^{\mathrm{tax}}_t$, providing an explicit, visually grounded signal for both impending and active contact. See Appendix B1 for implementation details.

\paragraph{Encoding multimodal observations}
We encode $\obscc_t$ into a latent $\lat_t$ using modality-specific encoders. We use a DINOv3 backbone \cite{simeoni2025dinov3} and apply it to $\rgb_t$, $\depth_t$, and $\proxmask_t$ using a shared encoder in a batched forward pass, with $\depth_t$ and $\proxmask_t$ replicated to three channels. DINOv3's robust spatial features transfer well to depth and mask inputs. For force sensing, we treat $\forces_t$ as a vector of normal forces under a fixed taxel ordering along the robot arm. Contact events in our dataset are spatially sparse. To efficiently detect these local features without significantly overparameterizing the model, we use a lightweight convolutional network with 1D \texttt{CoordConv} \cite{liu2018intriguing} layers. By concatenating the taxel index as an explicit position channel, this architecture enables the network to learn location-dependent contact features. For proprioception, we use a lightweight MLP that jointly embeds $(\q_t,\qd_t)$. To avoid angle wrap-around, we represent each joint angle using a sin-cos mapping before passing to the proprioceptive encoder. The latent dynamics model (Sec.~\ref{sec:hybrid_predictive_model}) conditions on a history of $\histlen$ latent states, $\lat_{t-\histlen+1:t}$. \looseness=-1

\subsection{Contact-Aware Action Sampling}
\label{sec:contact_aware_sampling}

We use MPPI, a sampling-based MPC algorithm, to generate command actions for the robot. At each MPC time step $t$, MPPI samples $\Nroll$ action sequences over a horizon of length $\horizon$. Standard MPPI samples perturbations $\pert^{(j)}_{t+i}\sim\mathcal{N}(\vct{0},\covAct)$ for rollout $j\in\{1,\ldots,\Nroll\}$ and horizon index $i\in\{0,\ldots,\horizon-1\}$, and sample action sequence $\act^{(j)}_{t:t+\horizon-1}$. We build on STORM~\cite{bhardwaj2022storm} as our base MPPI sampler, which uses low-discrepancy Halton sequences and spline-based sample smoothing, and adapts the mean and covariance parameters via cost-weighted sample moments.

In whole-arm manipulation, the set of active contacts can change from step to step. If contact handling
enters only through the scalar rollout cost, the sampler can adapt slowly: covariance updates lag behind
the current contact interaction, especially with a smaller particle budget. We therefore incorporate contact information directly into the sampler by shaping perturbations per-MPC-step, providing minimal-lag, contact-conditioned exploration toward the \emph{contact normal} subspace.

Although STORM is formulated over joint accelerations, we use it as an \emph{auxiliary sampler} to generate smooth perturbation sequences while keeping our MPPI policy in joint-velocity space. This allows for smoother perturbations in joint-velocity space. Concretely, for each rollout $j$, we draw acceleration noise $\accNoise^{(j)}_{t:t+\horizon-1}$ using Halton sampling with B-spline smoothing and integrate it over the horizon to obtain joint-velocity perturbations $\pert^{(j)}_{t:t+\horizon-1}$ (with time step $\dt$),
\(
    \pert^{(j)}_{t+i} = \dt \sum_{\ell=0}^{i}\accNoise^{(j)}_{t+\ell},
\)
and apply $\act^{(j)}_{t+i}=\nomAct_{t+i}+\pert^{(j)}_{t+i}$ to generate joint velocity action $\act^{(j)}_{t+i}$ around nominal mean $\nomAct_{t+i}$.
\paragraph{Weighted normal contact Jacobian} Let $\cset_t$ be the active contact set at time $t$. For each $k\in\cset_t$, we compute the contact
Jacobian $\mathbf{J}_{k,t}\in\R^{3\times \ndof}$ and a unit contact normal $\nrm_{k,t}\in\R^3$. The
corresponding normal Jacobian row \(\Jn_{k,t}=\nrm_{k,t}^\top \mathbf{J}_{k,t}\)
maps joint velocities to normal velocity at contact. Stacking rows gives $\Jn_t\in\R^{K_t\times \ndof}$, and the predicted normal velocity under rollout $j$ is $v^{(j)}_{n,k,t+i}=\Jn_{k,t}\act^{(j)}_{t+i}$. Intuitively, directions in joint-velocity space that lie in $\mathrm{row}(\Jn_t)$ change normal motion at the current contacts; directions in its nullspace do not.

Not all ``active contacts'' are equally important. A light brush or noisy taxel should not dominate sampling. Thus, we weight each contact by its measured force (with saturation) before constructing the contact-normal Jacobian:
\begin{equation}
    \contactw_{k,t} = \mathrm{sat}\!\left(\frac{\|\fmag_{k,t}\|}{\Fsat}\right), \quad
    \widetilde{\Jn_t}
    =
    \mathrm{diag}(\sqrt{\contactw_{1,t}},\ldots,\sqrt{\contactw_{K_t,t}})\,\Jn_t,
\end{equation}
where $\mathrm{sat}(x)=\min(1,\max(0,x))$. The resulting
$\widetilde{\Jn_t}$ maps joint velocities to normal contact velocities, and its row space spans the joint
directions that can change normal motion at the current contacts. 

\paragraph{Contact-conditioned shaping}
We define a contact-normal filter:

\begin{equation}
\begin{aligned}
\Pforce_t
&= \widetilde{\Jn_t}^{\top}
\left(\widetilde{\Jn_t}\widetilde{\Jn_t}^{\top} + \ridge\,\mathbf{I}_{K_t}\right)^{-1}
\widetilde{\Jn_t}, \\
\Pnull_t
&= \mathbf{I}_{\ndof} - \Pforce_t .
\end{aligned}
\end{equation}

$\Pforce_t$ extracts the component of a perturbation aligned with $\mathrm{row}(\widetilde{\Jn_t})$, the directions that change normal velocity at active contacts, while $\Pnull_t$ retains the complementary component. The ridge term $\ridge\,\mathbf{I}_{K_t}$ stabilizes the inverse when contact normals are nearly collinear. As $\ridge \to 0$, $\Pforce_t$ recovers the orthogonal projector onto $\mathrm{row}(\widetilde{\Jn_t})$ and $\Pnull_t$ recovers the projector onto its nullspace.

Assuming the contact configuration remains valid over the short planning horizon, we shape the perturbation
sequences using the above operators. Let $\pert^{(j)}_{t+i}$ denote the base perturbation. We construct
\begin{equation}
\begin{aligned}
\shapeMat_t
&=
\sigma_{\mathrm{force,eff}}\,\Pforce_t
+
\sigma_{\mathrm{null,eff}}\,\Pnull_t, \\
\sigma_{\bullet,\mathrm{eff}}
&=
\sqrt{\sigma_\bullet^2+\sigma_{\min}^2}, \quad
\bullet \in \{\text{force},\text{null}\}, 
\end{aligned}
\end{equation}
$\shapeMat_t$ is a contact-conditioned scaling that amplifies perturbations along contact-normal directions relative to nullspace directions. We apply it to obtain contact-conditioned perturbations,
\(
\pert^{(j),\mathrm{shaped}}_{t+i}
= \shapeMat_t\, \pert^{(j)}_{t+i}.
\)
Setting $\sigma_{\mathrm{force}} > \sigma_{\mathrm{null}}$ biases exploration toward directions that affect contact interaction; $\sigma_{\min}$ ensures a nonzero scale in both subspaces.

Finally, to avoid sampling actions that drive motion into contact where forces are high, we project each sampled action sequence onto a safety cone defined by the active normals above a force threshold. Let
$\cset^{\mathrm{viol}}_t=\{k\in\cset_t \mid \fmag_{k,t}\ge \Fsafe\}$ and let
$\mathbf{J}^{\mathrm{viol}}_{t}$ stack rows $\Jn_{k,t}$ for $k\in\cset^{\mathrm{viol}}_t$. For each rollout $j$ and horizon step $i$, we solve
\begin{equation}
    \pert^{(j),\mathrm{final}}_{t+i} =
    \argmin_{\vct{y}\in\R^{\ndof}}\norm{\vct{y}-\pert^{(j),\mathrm{shaped}}_{t+i}}_2^2
    \label{eq:contact_qp}
\end{equation}
$\text{s.t.}\quad
    \mathbf{J}^{\mathrm{viol}}_{t}\bigl(\nomAct_{t+i}+\vct{y}\bigr)\ge \vct{0},$
where rows are oriented such that $\mathbf{J}^{\mathrm{viol}}_{t+i}\act_{t}\ge \vct{0}$
corresponds to commanded normal motion away from contact. The obtained action is thus
$\act^{(j)}_{t+i}=\nomAct_{t+i}+\pert^{(j),\mathrm{final}}_{t+i}$. See Appendix C for implementation details.

\subsection{Contact-Centric Hybrid Predictive Model}
\label{sec:hybrid_predictive_model}

Whole-arm manipulation closely couples joint-space motion and contact: small changes in arm configuration can shift where contact occurs and whether forces increase or decrease. To plan under this coupling, TACTIC uses a \emph{hybrid predictive model} $\Fhyb=\{\Fr,\Fl\}$ that separates what we can compute reliably (robot kinematics) from what is difficult to model analytically (contact formation, making and breaking of contacts, and interaction dynamics). The analytical branch provides physically grounded rollouts, while the learned branch predicts the evolution of complex contact-rich interactions from multimodal observations.
\paragraph{Analytical kinematics $\Fr$}
Given the current joint state and a candidate joint-velocity sequence $\act_{t:t+\horizon-1}$, we roll
out deterministic kinematics with first-order integration to obtain $\hat{\q}_{t+i}, \hat{\qd}_{t+i}$ and compute task-space quantities via forward kinematics, $\hat{\vct{e}}_{t+i}=\mathrm{FK}(\hat{\q}_{t+i})$,
where $\hat{\vct{e}}_{t+i}$ denotes the predicted end-effector pose (or task-space state) at step $t+i$.
For contact-aware reasoning, we evaluate the contact Jacobian $\mathbf{J}_{k,t+i}(\hat{\q}_{t+i})$ for each current active contact index $k\in\cset_t$, at the predicted configuration $\hat{\q}_{t+i}$ and associate it with the contact normal $\nrm_{k,t+i}$ yielding the normal Jacobian row $\hat{\mathbf{J}}^{n}_{k,t+i}=\nrm_{k,t}^\top \hat{\mathbf{J}}_{k,t+i}$.
We collect these predictions into the kinematic rollout
\(
    \hat{s}_{t+i}
    \;=\;
    \Big(\hat{\q}_{t+i},\; \hat{\qd}_{t+i},\; \hat{\vct{e}}_{t+i},\; \{\nrm_{k,t},\,\hat{\mathbf{J}}_{k,t+i}\}_{k\in\cset_t}\Big).
\)
The kinematic rollout $\hat{s}_{t+1:t+\horizon}=\Fr(\q_t,\act_{t:t+\horizon-1})$ provides information needed to (i) reject kinematically infeasible motion and (ii) reason about how candidate actions move the active contacts along the contact normal (via $\hat{\mathbf{J}}^{n}_{k,t+i}$).

\paragraph{Latent dynamics $\mathbf{F}_l$}
Kinematics alone cannot predict when contacts will form or break. It also discards the multimodal context needed to perform the task. We therefore learn a latent dynamics model following the JEPA recipe for robotic world models~\cite{zhou2024dino, assran2025v}. At each time step, $\obscc_t$ is encoded into a latent state $z_t$ using the encoders in Sec.~\ref{sec:contact_centric_repr}. We embed $\act_t$ with a lightweight action encoder $\phi_{\mathrm{act}}(\act_t)$.
The predictor takes a context window $\lat_{t-\histlen+1:t}$ and an action sequence $\act_{t:t+\horizon-1}$ and outputs a latent rollout:
\(
    \hat{z}_{t+1:t+H}=\mathbf{F}_l\!\left(\lat_{t-\histlen+1:t},\,\phi_{\mathrm{act}}(\act_{t:t+H-1})\right).
\)
We implement $\mathbf{F}_l$ as a ViT \cite{dosovitskiy2020image} predictor with a frame-causal attention mask ~\cite{zhou2024dino}. We condition on actions using adaptive layer normalization at each transformer block, following best practices in~\cite{terver2025drives}. We use lightweight decoders $\psi_m$ to map latent rollouts to task-relevant signals. In this work, we decode joint position and binary contacts for all taxels. These decoders are used for force-based costs in Sec.~\ref{sec:hybrid_cost}.

\paragraph{Training objective}
We train our models on offline datasets. From each trajectory we sample segments containing a
context of length $\histlen$ and a prediction horizon $\horizon$, encode $\lat_t=\enc(\obscc_t)$, and predict
$\latHat_{t+1:t+\horizon}=\Fl(\lat_{t-\histlen+1:t},\,\phi_{\mathrm{act}}(\act_{t:t+\horizon-1}))$.
Based on findings shared in \cite{terver2025drives}, we minimize a teacher-forced latent consistency loss, augmented with a multistep rollout loss (to reduce drift when the model must consume its own predictions):
\begin{equation*}
\mathcal{L}
=
\underbrace{\sum_{k=1}^{\horizon}\bigl\|\latHat_{t+k}-\mathrm{sg}[\lat_{t+k}]\bigr\|_2^2}_{\mathcal{L}_{\mathrm{teacher}}}
+
\underbrace{\sum_{k=2}^{K_{\mathrm{roll}}}\lambda_k\,\bigl\|\latHat^{\langle k\rangle}_{t+k}-\mathrm{sg}[\lat_{t+k}]\bigr\|_2^2}_{\mathcal{L}_{\mathrm{multistep}}}
\end{equation*}
where $\latHat^{\langle k\rangle}_{t+k}$ denotes the $k$-step autoregressive prediction (up to $K_{roll}$) obtained by feeding back the model’s own latent predictions, and $\mathrm{sg}[\cdot]$ is stop-gradient on the target latent.

\paragraph{Planning with hybrid models}
At each MPC step $t$, we form the contact-centric observation $\obscc_t$ (Sec.~\ref{sec:contact_centric_repr}) and use MPPI with contact-aware action sampling (Sec.~\ref{sec:contact_aware_sampling}) to sample $\Nroll$ candidate action sequences $\act^{(j)}_{t:t+\horizon-1}$, $j=1,\ldots,\Nroll$. For each candidate, we roll out two models in parallel. The analytical branch propagates joint kinematics via
$\hat{\q}^{(j)}_{t+1:t+\horizon}=\Fr(\q_t,\act^{(j)}_{t:t+\horizon-1})$ and produces kinematic rollout
$\hat{s}^{(j)}_{t+1:t+\horizon}$ (combining joint states, end-effector states, contact Jacobians for active contacts). Meanwhile, we encode a history of observations into a latent context $\lat_{t-\histlen+1:t}=\enc(\obscc_{t-\histlen+1:t})$ and unroll the learned latent dynamics
$\latHat^{(j)}_{t+1:t+\horizon}=\Fl(\lat_{t-\histlen+1:t},\act^{(j)}_{t:t+\horizon-1})$ (Sec.~\ref{sec:hybrid_predictive_model}). The resulting rollouts are scored by $\Cost(\act^{(j)}_{t:t+\horizon-1})$, a weighted combination of different cost functions (discussed in Sec.~\ref{sec:hybrid_cost}). MPPI then updates the nominal sequence toward low-cost rollouts and returns $\act^\star_{t:t+\horizon-1}$; we execute the first $m$ actions $\act^\star_{t:t+m-1}$ and replan at $t+m$ with updated observations.

\paragraph{Two-fidelity rollouts for real-time control}
Evaluating $\mathbf{F}_l$ for all $N$ MPPI particles can dominate compute and slow down planning significantly. We use a two-fidelity scheme: a
lightweight \emph{draft} predictor $\mathbf{F}_l^{\mathrm{draft}}$ rolls out all particles to obtain approximate costs, then only the top-$M$ candidates ($M\ll N$) are re-scored with the full predictor $\mathbf{F}_l$ before computing MPPI weights. This reduces large ViT evaluations while preserving ranking quality on the best candidates and enabling real-time control. As a safeguard, we monitor the one-step prediction error
$\|\hat{z}^{\mathrm{draft}}_{t+1}-z_{t+1}\|$ and fall back to the full model when the error exceeds a threshold
(see Appendix D). \looseness=-1
\subsection{Scoring Hybrid Rollouts}
\label{sec:hybrid_cost}
We obtain the best action sequence out of the $\Nroll$ sampled sequences by scoring the rollouts generated by our hybrid dynamics model using our cost function $\Cost$ defined as:
\vspace{-1mm}
\begin{align*}
\label{eq:hybrid_cost}
\Cost(\act_{t:t+\horizon-1})
&= \sum_{i=1}^{\horizon} \Big(
w_{\text{kin}}\,\ckin(\hat{s}_{t+i}) \\
&\qquad
+ w_{\text{safety}}\,
\cforce(\hat{s}_{t+i}, \latHat_{t+i}, \forces_t) \\
&\qquad
+ w_{\text{goal}}\,\cgoal(\latHat_{t+i}, \latGoal)
\Big),
\end{align*}
where $\ckin$ penalizes for standard kinematic constraints (self-collision, joint-limit, joint or task-space goals),
$\cforce$ penalizes unsafe predicted forces relative to $\Fsafe$, combining the learned contact-onset predictor with an analytical force model, and $\cgoal(\latHat_{t+i},\latGoal)$ encourages goal-reaching in latent space. We provide the exact cost definitions used in our experiments in Appendix E.

\textbf{Learned value function for long-horizon tasks}
To improve latent-space planning on constrained long-horizon tasks, we learn a goal-conditioned value
function in latent space from offline data using Implicit Q-Learning (IQL)~\cite{kostrikov2021offline}. For such long-horizon tasks, we assume access to low-dimensional progress variables $\xi_t\in\mathbb{R}^{d_g}$ provided by the dataset
(e.g., limb joint angles or body turn angle). At test time, we use a user-specified sequence of \emph{subgoals}
$\{\xi_g^{(1)},\ldots,\xi_g^{(G)}\}$, where each $\xi_g^{(g)}$ represents a desired intermediate progress configuration.

We freeze the
multimodal encoders (Sec.~\ref{sec:contact_centric_repr}) and embed observations and (sub)goals into latents
$\lat_t$ and $\latGoal$. At test time, we augment MPPI with a terminal value term evaluated at the predicted
end-of-horizon latent state:
\begin{align*}
\Cost_{\mathrm{aug}}(\act_{t:t+\horizon-1})
=
\Cost(\act_{t:t+\horizon-1})
\;-\;
\wVal\,\Vfn(\latHat_{t+\horizon}, \latGoal),
\end{align*}
where $\latHat_{t+\horizon}$ is obtained from $\Fl$. Intuitively,
$\Vfn$ estimates reward-to-go beyond the finite horizon, biasing sampling toward action sequences that
remain on-task over extended horizons while preserving kinematic feasibility and safety induced by $\ckin$ and $\cforce$. Training and
implementation details are provided in Appendix F.

\vspace{-1mm}
\section{Simulation Experiments}
\label{sec:sim_experiments}

We first evaluate TACTIC in simulation to isolate the benefits of our contributions by posing the questions below:
\textbf{Q1:} does having contact-centric representations \textbf{(C1)} comprising distributed tactile sensing and the proximity mask improve task success and safety?
\textbf{Q2:} does contact-aware action sampling \textbf{(C2)} improve interaction response, safety, and smoothness?
\textbf{Q3:} does hybrid modeling \textbf{(C3)} by coupling analytical kinematics with learned latent dynamics improve success and efficiency versus either alone?

\begin{figure*}[t]
    \centering
    \includegraphics[width=\textwidth]{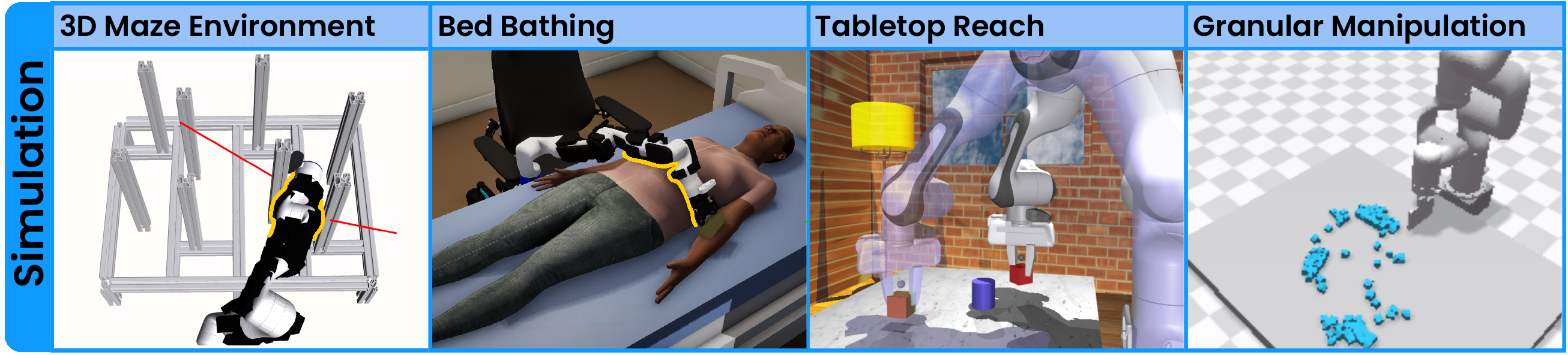}
    \caption{\textbf{Simulation task suite.} We evaluate TACTIC across four simulation environments: the \textbf{3D Maze Environment} tests whole-arm navigation through a maze, analogous to navigation in a cluttered environment; \textbf{Bed Bathing} tests task execution of wiping a person in bed with incidental whole-arm contact; \textbf{Tabletop Reach} tests obstacle-aware reaching across a table; and \textbf{Granular Manipulation} tests contact-driven rearrangement of small particles. Whole-arm contact is highlighted in yellow.}

    \label{fig:sim_tasks}
\end{figure*}

\noindent\textbf{Environments.} We evaluate on 4 environments (Fig. \ref{fig:sim_tasks}):

\begin{itemize}[leftmargin=*]
\item\textbf{3D Maze Goal Reaching (Maze)} \textit{RCareWorld} \cite{ye2022rcare}.
A 3D maze composed of cylinders placed at varying $x$--$y$ coordinates. Collision-free motion is often infeasible; successful solutions require deliberate whole-arm contact to slide past obstacles and reach a target end-effector pose. We define force thresholds and assess safety via threshold violations.

\item\textbf{Bed Bathing (Wiping)} \textit{RCareWorld}.
A wheelchair-mounted arm performs a sponge-wiping task on a user lying in a hospital bed. The end-effector trajectory is constrained to follow a wiping motion along the user’s arm. Incidental whole-arm contact occurs as the robot reaches over the bed.

\item\textbf{Tabletop Reach (Reach)} \textit{Robohive} \cite{kumar2023robohive, sorensen2026robohive}.
A Franka arm reaches across a table to move a cube to a target while avoiding obstacles. This task is used to evaluate hybrid modeling with pretrained world models.

\item\textbf{Granular Manipulation (Granular)} \textit{FleX} \cite{zhang2024adaptigraph}.
An xArm manipulates particles on a tabletop and must transform an initial distribution into a target pile configuration. We use this task to evaluate hybrid modeling with a pretrained granular world model.
\end{itemize}

\noindent \textbf{Data and Training Protocol.} We collect whole-arm manipulation data for \textbf{Maze}, \textbf{Bathing}, and \textbf{Reach} using (i) RRT-generated feasible start/goal configurations and (ii) random-play trajectories obtained by applying noisy joint-velocity commands from feasible states. For \textbf{Granular}, we use the dataset released by~\cite{zhou2024dino}.
All learning-based components (world models, value functions, and baselines) are trained \emph{offline} on these datasets; no method uses online environment interaction for training.

\textbf{Ablations (A1-4) and Baselines (B1-4).}
We perform all ablations in \textbf{Maze} to minimize confounds such as:
\textbf{A1.} TACTIC-NoTacNoMask (Q1), which removes distributed tactile readings and the proximity mask, using RGB-D and proprioception only. 
\textbf{A2.} TACTIC-NoTac (Q1), which removes tactile readings $\forces_t$ while retaining the vision-based proximity mask, isolating the effect of the mask when tactile information is unavailable. 
\textbf{A3.} TACTIC-NoCaSa (Q2), which replaces contact-aware sampling with the base STORM sampler~\cite{bhardwaj2022storm}.
\textbf{A4.} TACTIC-L / TACTIC-K / TACTIC (Q3), which evaluates learned latent dynamics only, kinematics only, or the full hybrid rollout model (Sec.~\ref{sec:hybrid_predictive_model}).

\textbf{Hybridization with pretrained world models.}
Given the general applicability of hybrid modeling, we evaluate it on manipulation tasks that extend beyond whole-arm manipulation. Specifically, we run experiments with pretrained world models: \textbf{B1.} V-JEPA2 \cite{assran2025v} on Reach, and \textbf{B2.} DINO-WM \cite{zhou2024dino} on Granular. In both cases, we add a kinematics-based goal cost to the planner to incorporate analytical structure on top of the learned predictor, while keeping the pretrained encoder/predictor intact unless finetuning is explicitly stated.

\textbf{Model-based baselines.}
We also compare TACTIC against two offline model-based RL baselines: \textbf{B3.} DreamerV3~\cite{hafner2023mastering} is a strong world-model RL method that learns a latent dynamics model together with actor/value functions. This provides a relevant comparison because it represents an alternative end-to-end way to act from a learned model––learned rollout and value-based policy––without explicit contact-aware sampling or hybrid analytical rollouts. \textbf{B4.} TD-MPC2~\cite{hansen2024tdmpc2} is a competitive, decoder-free model-based method that learns latent dynamics and value estimates and can plan in latent space.

\textbf{Metrics.} We use the following metrics:
\begin{itemize}
    \item \textbf{Success rate (SR)}: \% age of trials completing the task before timeout.
    \item \textbf{Time to completion (TTC)}: time (in s) to success.
    \item \textbf{Force violations (FV)}: count of unsafe-force timesteps
\end{itemize}

\textbf{Results.} \textbf{Contact-centric sensing improves safety and performance.}
We ablate tactile sensing and proximity mask in \textbf{Maze}.
Removing both degrades performance and increases unsafe interactions.

\begin{table}[t]
\centering
\footnotesize
\setlength{\tabcolsep}{6pt}
\renewcommand{\arraystretch}{1.15}
\begin{tabularx}{\linewidth}{l *{3}{>{\centering\arraybackslash}X}}
\toprule
\textbf{Variant} & \textbf{SR} $\uparrow$ & \textbf{TTC} $\downarrow$ & \textbf{FV} $\downarrow$ \\
\midrule
TACTIC-NoTacNoMask & 63.5 & \textbf{20.3 $\pm$ 4.7} & 84.4 $\pm$ 53.9 \\
TACTIC-NoTac       & 72.1 & 23.8 $\pm$ 8.5 & 55.1 $\pm$ 56.7\\
\textbf{TACTIC}    & \textbf{87.2} & 22.4 $\pm$ 8.1 & \textbf{39.0 $\pm$ 55.9} \\
\bottomrule
\end{tabularx}
\vspace{2pt}
\caption{Contact-centric representation boosts task performance.}
\label{tab:modality_ablation}
\end{table}

\textbf{Contact-aware sampling reduces unsafe interactions and improves smoothness.}
Tab.~\ref{tab:contact_aware_results} compares TACTIC against a variant that uses the base STORM sampler,
keeping the predictive model and costs fixed.
Injecting local contact geometry into the sampler yields substantially fewer force violations and smoother motion (lower jerk and action magnitude).

\begin{figure*}[t]
    \centering
    \includegraphics[width=\textwidth]{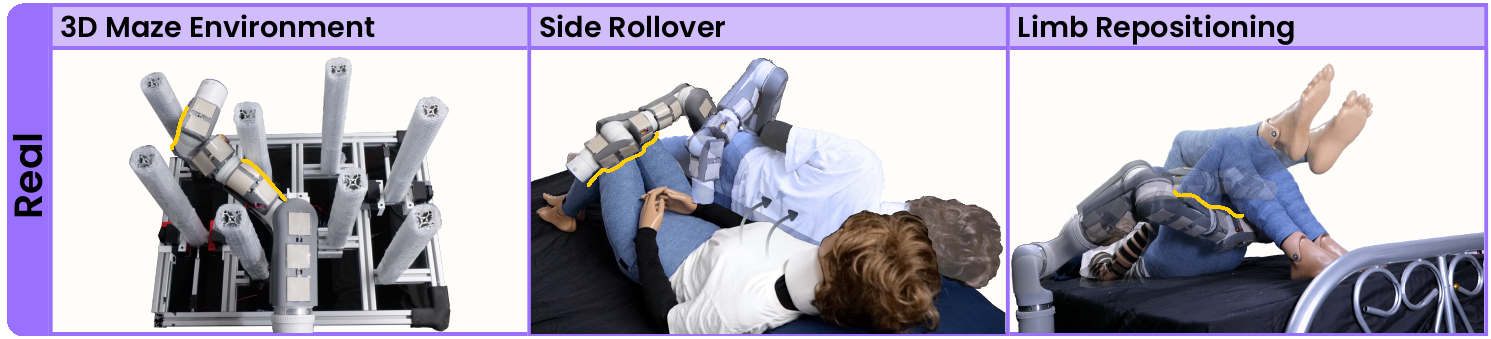}
    \caption{\textbf{Real-world task suite.} We evaluate across three real-world environments. The \textbf{3D Maze Environment} tests whole-arm navigation through static and dynamic obstacles, \textbf{Side Rollover} tests contact-rich manipulation to rotate a person while regulating interaction forces, and \textbf{Limb Repositioning} tests multi-contact support during lifting and repositioning of the legs to a target configuration. Whole-arm contact is highlighted in yellow.}
    \label{fig:real_tasks}
\end{figure*}

\begin{table}[t]
\centering
\footnotesize
\setlength{\tabcolsep}{6pt}
\renewcommand{\arraystretch}{1.18}
\begin{tabularx}{\linewidth}{l *{4}{>{\centering\arraybackslash}X}}
\toprule
\textbf{Method} &
\textbf{FV} $\downarrow$ &
\textbf{Mean Jerk} $\downarrow$ \\
\midrule
TACTIC-NoCaSa & $111.9 \pm 71.2$ &  $5.87 \pm 2.98$ \\
\textbf{TACTIC} & \textbf{39.0 $\pm$ 55.9}  &  \textbf{2.83 $\pm$ 3.23}  \\
\bottomrule
\end{tabularx}
\caption{Contact-aware sampling improves safety and smoothness.}
\label{tab:contact_aware_results}
\end{table}

\textbf{Hybrid rollouts improve success and reduce unsafe contact.}
Tab.~\ref{tab:hybrid_success_sampleeff} compares latent-only, kinematics-only, and hybrid rollouts.
Kinematics-only rollouts provide geometric grounding but cannot anticipate contact state changes.
Latent-only rollouts capture interaction patterns but can drift without analytical structure.
Hybrid rollouts achieve the best success/safety trade-off.

\begin{table}[t]
\centering
\footnotesize
\setlength{\tabcolsep}{6pt}
\renewcommand{\arraystretch}{1.15}
\begin{tabularx}{\linewidth}{l *{3}{>{\centering\arraybackslash}X}}
\toprule
\textbf{Variant} & \textbf{SR} $\uparrow$ & \textbf{TTC} $\downarrow$ & \textbf{FV} $\downarrow$ \\
\midrule
TACTIC-L & $64.1$ & $47.9 \pm 13.4$& 108.3 $\pm$ 41.4 \\ 
TACTIC-K & $43.6$ & $71.8 \pm  4.3$& 79.9 $\pm$ 38.2\\
\textbf{TACTIC} & \textbf{87.2} & \textbf{22.4 $\pm$ 8.1} & \textbf{39.0 $\pm$ 55.9} \\ 
\bottomrule
\end{tabularx}
\caption{TACTIC outperforms latent- and kinematic-only variants.}
\label{tab:hybrid_success_sampleeff}
\end{table}

\looseness=-1
\textbf{Hybrid modeling benefits pretrained world-model planning.} We study two pretrained predictors: \textbf{V-JEPA2} on \textbf{Reach} and \textbf{DINO-WM} on \textbf{Granular}. In both cases, we compare the original planner to a \emph{hybrid} variant that adds a kinematics-based goal cost during planning, keeping the pretrained encoder/predictor frozen unless finetuning is explicitly stated. Hybridization improves V-JEPA2 finetuning data efficiency (achieving comparable performance with $\sim$34\% less finetuning data) and improves geometric rollout quality for frozen DINO-WM (reducing Chamfer Distance by $\sim$26\%; see Appendix I). 

\looseness=-1
\textbf{TACTIC outperforms alternate model-based approaches.}
Tab.~\ref{tab:sim_baselines} shows TACTIC achieves the highest success and lowest force violations on \textbf{Maze} compared to DreamerV3 and TD-MPC2, and substantially improves success and safety on \textbf{Bed Bathing} compared to offline DreamerV3.

\begin{table}[t]
\centering
\scriptsize
\setlength{\tabcolsep}{3pt}
\renewcommand{\arraystretch}{1.0}
\begin{tabularx}{\columnwidth}{l *{4}{>{\centering\arraybackslash}X}}
\toprule
\multirow{2}{*}{\textbf{Method}} &
\multicolumn{2}{c}{\textbf{Maze}} &
\multicolumn{2}{c}{\textbf{Bed Bathing}} \\
\cmidrule(lr){2-3}\cmidrule(lr){4-5}
& \textbf{SR} $\uparrow$ & \textbf{FV} $\downarrow$
& \textbf{SR} $\uparrow$ & \textbf{FV} $\downarrow$ \\
\midrule
TD-MPC2~\cite{hansen2024tdmpc2} & 65.1 & 93.9 $\pm$ 34.0 & -- & -- \\
DreamerV3~\cite{hafner2023mastering} & 75.9 & 82.4 $\pm$ 28.7 & 51.6 & 63.8 $\pm$ 17.4 \\
\textbf{TACTIC (ours)} & \textbf{87.2} & \textbf{39.0 $\pm$ 55.9} & \textbf{79.5} & \textbf{22.6 $\pm$ 45.2} \\
\bottomrule
\end{tabularx}
\vspace{1pt}
\caption{TACTIC outperforms alternate model-based methods.}
\label{tab:sim_baselines}
\end{table}

\section{Real World Experiments}
\label{sec:real_experiments}
While simulation validates our design choices in controlled settings, real-world deployment requires real-time control under sensing noise, actuation delays, and changing contact conditions. We evaluate TACTIC on a real robot in (i) a dynamic cluttered maze that tests contact-rich navigation involving highly constrained multi-contact trajectories, and (ii) contact-intensive manipulation tasks with a life-size manikin ($\sim$46 lbs, vinyl plastic) of the type used for nurse training on the tasks studied here, that require multi-contact regulation and goal-directed motion under safety limits. These experiments assess whether the contact-aware planning and hybrid modeling principles validated in simulation carry over to realistic manipulation scenarios.
\begin{figure}[t]
    \centering
    \includegraphics[width=\linewidth]{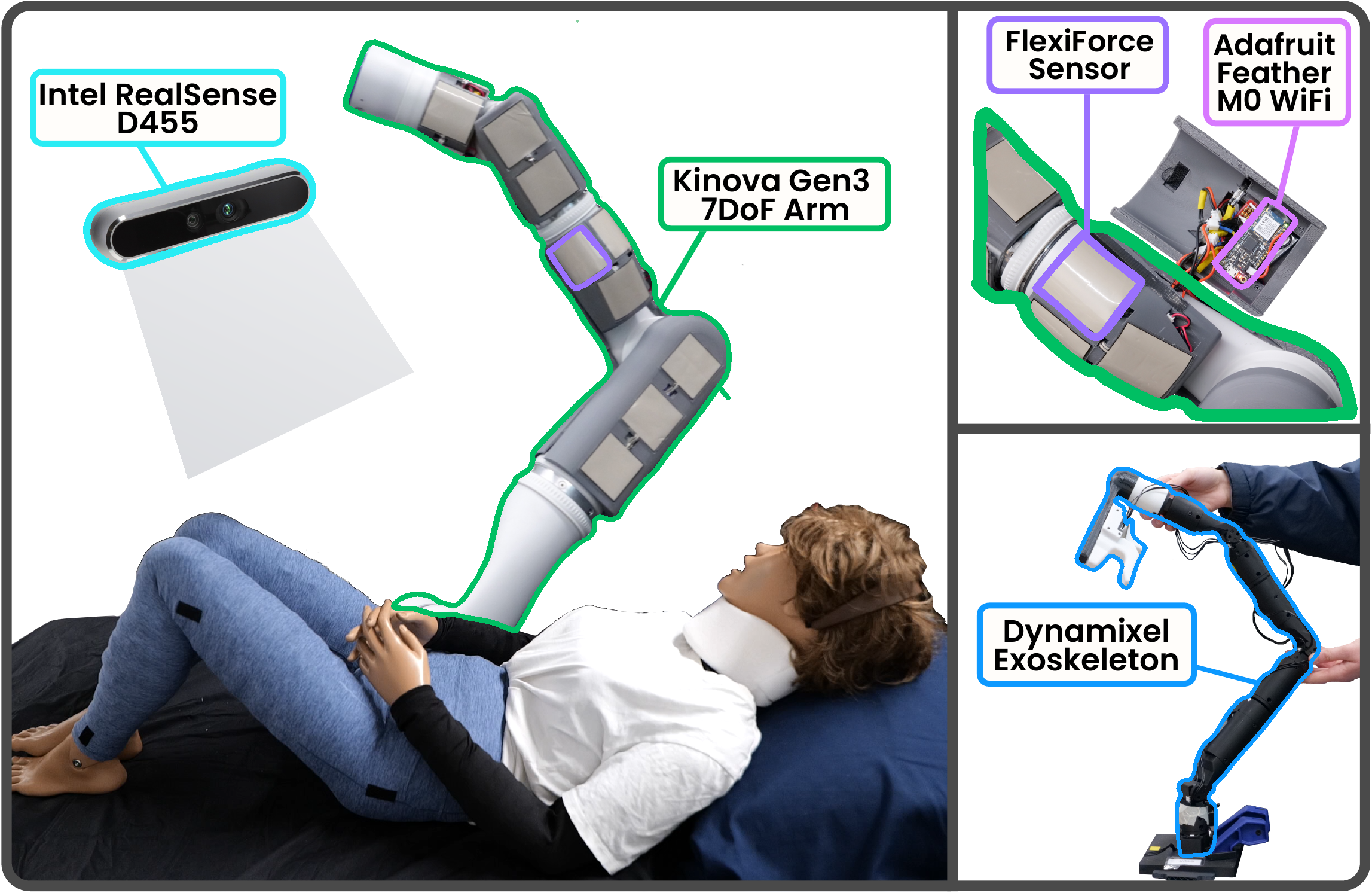}
    \caption{\textbf{Real-world setup.} Kinova Gen3 7-DoF arm with 22 distributed tactile sensors, RGB-D perception, and exoskeleton-based teleoperation for data collection.}
    \label{fig:real_setup}
\end{figure}

\textbf{Environments.} We evaluate on 3 environments (Fig. \ref{fig:real_tasks}):

\begin{itemize}[leftmargin=*]
    \item \textbf{3D maze environment.}
    Real-world instantiation of the maze task. Four central cylinders translate with amplitude 3\,cm and speed 1\,cm/s. The robot must reach a target pose while safely making contact when necessary.
    \item \textbf{Side rollover.}
    The robot rolls a manikin from a back-lying position onto its side while regulating contact forces to remain within safe limits.
    \item \textbf{Limb repositioning (LimbRepo).}
    The robot lifts and repositions the manikin’s legs to a target configuration (e.g., target joint angle or clinician-provided range-of-motion goal), often requiring multi-contact support across motion phases.
\end{itemize}

\textbf{Robot Setup and Data Collection.} We use a Kinova Gen3 7-DoF arm (Figure \ref{fig:real_setup}) with a novel tactile skin comprising 22 Tekscan FlexiForce sensors \cite{tekscan_flexiforce} mounted along the links.
Sensor readings are transmitted through four Adafruit Feather M0 WiFi boards at $\sim$800\,Hz per board.
RGB-D is captured by an overhead Intel RealSense D455.
We collect (i) expert demonstrations via our custom exoskeleton-based teleoperation interface and (ii) random-play trajectories for coverage. We open-source the exoskeleton and WiFi-based distributed piezoresistive tactile sensing suite (see \href{https://emprise.cs.cornell.edu/tactic}{website} for details). TACTIC is deployed as one layer of a multi-tier control stack. At the planning layer, an MPPI loop runs at $\sim$12 Hz. Below the planner, a joint-space compliant controller running at 1 kHz tracks $\act^\star_{t}$. Force thresholds, controller gains, the full set of safety-relevant constants are reported in Appendix C, and runtime statistics in Appendix A.

\textbf{Baselines}
The model-based RL baselines (DreamerV3, TD-MPC2) underperformed TACTIC in our simulation benchmarks, and their real-world performance would be difficult to interpret. We compare against an offline imitation baseline and report expert teleoperation as a safety reference:

\begin{tightlist}
    \item \textbf{Diffusion Policy (DP)~\cite{chi2023diffusion}.}
    A widely used imitation learning baseline for real-world robot manipulation. It is appropriate in our setting because it can be trained directly from a limited number of expert demonstrations.
    \item \textbf{Human expert teleoperation reference.}
    We pair each autonomous trial with a teleop episode collected under the same environment configuration (start/goal parameters) and ask whether TACTIC can achieve comparable task performance while outperforming the reference in terms of force violations.
\end{tightlist}

\textbf{Results.} Tab. \ref{tab:real_results} shows TACTIC achieves higher success than Diffusion Policy on two out of three tasks while being comparable on Maze. DP shows no violation for maze since it usually enters from the top of the maze and fits into the final configuration without going ``through'' the maze. TACTIC produces fewer force-threshold violations than matched expert teleoperation episodes, indicating that contact-aware planning can regulate forces beyond demonstration distribution and do better than the human reference.

\begin{table}[t]
\centering
\scriptsize
\setlength{\tabcolsep}{3pt}
\renewcommand{\arraystretch}{1.0}
\begin{tabularx}{\columnwidth}{l l *{3}{>{\centering\arraybackslash}X}}
\toprule
\textbf{Task} & \textbf{Method} &
\textbf{SR} $\uparrow$ &
\textbf{TTC} $\downarrow$ &
\textbf{FV} $\downarrow$ \\
\midrule
\multirow{3}{*}{Maze} 
& Expert Teleop & 5/5 & 30.4 $\pm$ 8.7 & 97.3 $\pm$ 50.1 \\
& DP~\cite{chi2023diffusion} & 2/5 & 300 $\pm$ 45.2 & 0.0 $\pm$ 0.0 \\
& TACTIC & 3/5 & 71.4 $\pm$ 10.4 & \textbf{70.3 $\pm$ 29.4} \\
\midrule
\multirow{3}{*}{Side Rollover} 
& Expert Teleop & 20/20 & 45.5 $\pm$ 13.6 & 141.2 $\pm$ 60.0 \\
& DP~\cite{chi2023diffusion} & 3/20 & 275 $\pm$ 43.1 & 150.5 $\pm$ 32.0 \\
& TACTIC & 12/20 & 81.0 $\pm$ 19.8 & \textbf{104.3 $\pm$ 46.5}\\
\midrule
\multirow{3}{*}{LimbRepo} 
& Expert Teleop & 20/20 & 30.4 $\pm$ 10.5 & 50.5 $\pm$ 24.0 \\
& DP~\cite{chi2023diffusion} & 5/20 & 262 $\pm$ 53.0  & 61.4 $\pm$ 22.8 \\
& TACTIC & 14/20 & 74.4 $\pm$ 8.2& \textbf{38.0 $\pm$ 16.5} \\
\bottomrule
\end{tabularx}
\caption{TACTIC outperforms DP on the Rollover and LimbRepo tasks. TACTIC also shows fewer force violations compared to the expert reference, i.e., the data it was trained on.}
\label{tab:real_results}
\end{table}

\vspace{-1mm}
\section{Discussion}

We presented TACTIC, a receding-horizon controller framework for whole-arm manipulation that embraces whole-arm contact through contact-centric representation, contact-aware sampling, and hybrid modeling. Our results confirm that jointly reasoning over analytical kinematics and learned latent dynamics outperforms pure learning approaches (DreamerV3, Diffusion Policy) in task performance and safety adherence. By shaping exploration via the contact Jacobian, TACTIC efficiently discovers valid sliding and pivoting motions.
TACTIC does not provide formal safety guarantees. The low-level compliant controller makes the interactions comfortable under arbitrary planner output, while TACTIC reduces unsafe contact forces at the planning stage. Both complement safer hardware, watchdogs, fallbacks, and emergency stops in a broader deployment-ready stack.
Future work can look into tighter integration with low-level control, where TACTIC also modulates stiffness and damping (variable impedance). Extending evaluation from manikins to human subjects is essential for validating both task performance and user comfort in caregiving settings. One scalability limitation of our framework is our reliance on a custom exoskeleton for high-fidelity data; we aim to reduce data dependence through cross-embodiment learning and investigate active tactile exploration to reduce state uncertainty.

\section*{Acknowledgements}
We would like to thank Ben Dodson and Alexios Rekoutis for the initial design of the exoskeleton used for data collection.  We would also like to thank Kapil Gangwar and Devesh Khilwani for help with debugging the circuit for the wireless sensor. This work was partly funded by National Science Foundation IIS \#2132846, and CAREER \#2238792. The Toyota Research Institute also partially supported this work. This article solely reflects the opinions and conclusions of its authors and not TRI or any other Toyota entity.

\bibliographystyle{unsrt}
\bibliography{references}

\clearpage
\onecolumn

\thispagestyle{empty}
\phantomsection
\label{sec:appendix-contents}

\vspace*{1.0em}
\hspace*{-1em}{\Large\scshape Table of Contents}
\vspace{1.25em}

\renewcommand{\contentsname}{}
\setcounter{tocdepth}{2}
\tableofcontents
\clearpage

\twocolumn

\setcounter{section}{0}
\setcounter{subsection}{0}
\setcounter{subsubsection}{0}
\setcounter{secnumdepth}{3}
\renewcommand{\thesection}{\Alph{section}}
\renewcommand{\thesubsection}{\arabic{subsection}}
\renewcommand{\thesubsubsection}{\thesubsection.\arabic{subsubsection}}
\renewcommand{\thesubsectiondis}{\arabic{subsection}.}
\renewcommand{\thesubsubsectiondis}{\arabic{subsection}.\arabic{subsubsection}.}
\renewcommand{\theHsection}{appendix.\Alph{section}}
\renewcommand{\theHsubsection}{appendix.\Alph{section}.\arabic{subsection}}
\renewcommand{\theHsubsubsection}{appendix.\Alph{section}.\arabic{subsection}.\arabic{subsubsection}}

\addtocontents{toc}{\protect\setcounter{tocdepth}{2}}
\addtocontents{toc}{\protect\apptocon}

\section{System Overview and Latency Breakdown}
\label{sec:app-overview}

\subsection{Pipeline overview}
\label{sec:app-overview-pipeline}

At each MPC step $t$, the system runs the following sequence:
\textbf{Sensor inputs} $\rightarrow$ \textbf{Mask generation} $\rightarrow$ \textbf{Contact-centric observation}
$\rightarrow$ \textbf{Encoding} $\rightarrow$ \textbf{Rollouts (draft + kinematic + full) + decoding}
$\rightarrow$ \textbf{Scoring} $\rightarrow$ \textbf{Execution}.
We acquire synchronized sensor data at \texttt{40}\,Hz. We then generate the proximity mask $\proxmask_t$ (segmentation + kd-tree queries + taxel projection) which takes $\sim$5--8\,ms. We assemble the contact-centric observation $\obscc_t$ and encode it into a latent
$\lat_t=\enc(\obscc_t)$, updating the rolling context $\lat_{t-\histlen+1:t}$ ($\sim$6ms).
After receiving the sensor observation, in parallel, MPPI samples $\Nroll$ action sequences and applies contact-aware shaping. For each candidate sequence, we roll out analytical kinematics $\Fr$ to obtain kinematic signals
$\hat{s}^{(j)}_{t+1:t+\horizon}$ and roll out the draft latent predictor
$\Fl^{\mathrm{draft}}$ for all particles to obtain predicted latents; these two rollouts are computed in parallel. We then select the top-$M$ candidates by draft
score and re-score only those with the full predictor, producing the final costs used to
compute MPPI weights and update the nominal sequence. Finally, we execute the first $m$
actions $\act^\star_{t:t+m-1}$ using a low-level controller running at \texttt{1000} Hz (joint-space compliant controller), and replan at $t+m$ with updated observations. Overall, our MPC runs at close to 12\,Hz.

\subsection{Hyperparameter table}
\label{sec:app-hparams}
\begin{table}[h]
  \centering
  \caption{Hyperparameters for TACTIC.}
  \label{tab:app-hparams}
  \begin{tabular}{ll}
    \toprule
    Parameter & Value \\
    \midrule
    MPC horizon $H$ & \texttt{8} \\
    MPC time step $\Delta t$ & \texttt{0.025} \\
    MPPI particles $N$ & \texttt{100} \\
    Two-fidelity top-$M$ & \texttt{10} \\
    Latent history length $L$ & \texttt{3} \\
    Control mode & \texttt{joint-space impedance} \\
    \bottomrule
  \end{tabular}
\end{table}

\section{Contact-Centric Representation Details}
\label{sec:app-contact-repr}

\subsection{Proximity mask construction}
\label{app:proxmask_details}

We assume RGB and depth are spatially aligned and obtain camera intrinsics $\mathbf{K}$ using the \texttt{pyrealsense} API. We use a fixed extrinsic transform between the robot base and camera, which is needed only for
projecting taxels into the image. We use the \texttt{easy\_handeye} \cite{easy_handeye} package to obtain the extrinsics. For each pixel $(u,v)$ with valid depth $z(u,v)$, we unproject to 3D in the camera frame as
$\vct{p}(u,v)=z\,\mathbf{K}^{-1}[u,\,v,\,1]^\top$.
We form $P_r=\{\vct{p}(u,v)\mid (u,v)\in\Omega_r\}$ and
$P_h=\{\vct{p}(u,v)\mid (u,v)\in\Omega_o\}$ from robot/object masks. The object masks are obtained using YOLO11n-seg. To train this model, we semi-automatically annotate the robot, maze, and manikin classes in our dataset using SAM2.1. After training, we convert the model to TensorRT and run the model with an inference time of 2--3\,ms. We build a kd-tree on $P_h$ and query each $\vct{p}\in P_r$ to compute
$d(\vct{p})=\min_{\vct{p}_h\in P_h}\|\vct{p}-\vct{p}_h\|_2$. We convert distances to an image-aligned proximity field by applying the following linear clipping:
\[
M^{\mathrm{vis}}_t(u,v)=\mathrm{clip}\!\left(1-\frac{d(\vct{p}(u,v))}{d_{\max}},\,0,\,1\right),
\]
where $d_{\max}$ is the maximum distance considered relevant for anticipating contact. Pixels outside
$\Omega_r$ are set to zero. We render a separate blank mask $\proxmask^{\mathrm{tax}}_t$ (zeros, same resolution as the image).
For each taxel $k$ with measured force exceeding $f_{\min}$, we compute its 3D pose via forward
kinematics, transform it into the camera frame, project it with $\mathbf{K}$, and draw a filled disk of
radius $r$ pixels centered at the projected location. We fuse the visual proximity field and taxel indicator by per-pixel maximum and clip to $[0,1]$:
\[
\proxmask_t(u,v)=\max\!\left(M^{\mathrm{vis}}_t(u,v),\,\proxmask^{\mathrm{tax}}_t(u,v)\right).
\]

In all experiments, we use $d_{\max}=\texttt{0.05}\,\mathrm{m}$, $f_{\min}=\texttt{2}\,\mathrm{N}$, and $r=\texttt{5}\,\mathrm{px}$.

\subsection{Encoder implementation details}
\label{app:encoder_details}

\begin{figure}[h]
    \centering
    \includegraphics[width=\columnwidth]{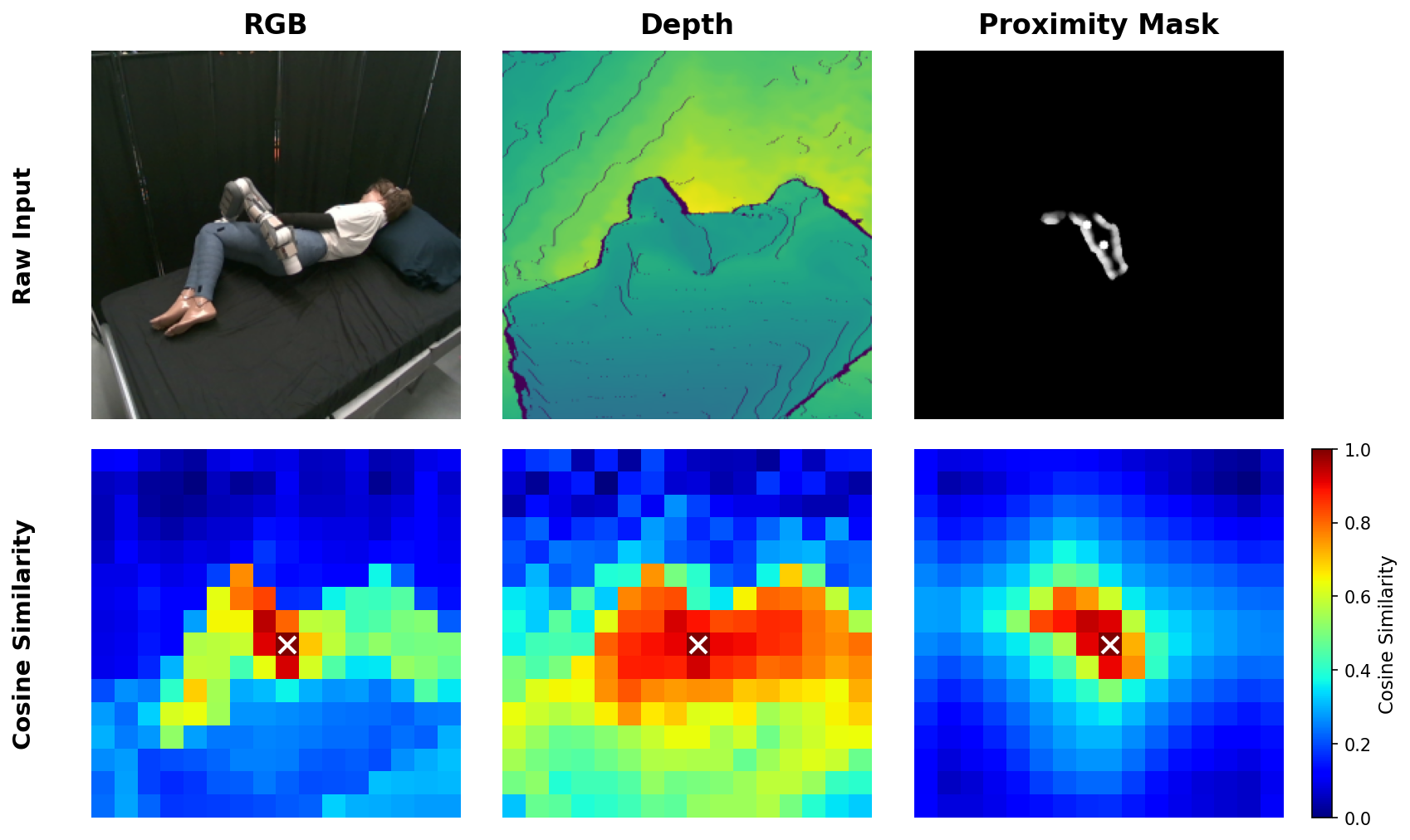}
    \caption{Cosine similarity for RGB, Depth, and Proximity Mask.}
    \label{fig:dino_vis}

\end{figure}

\textbf{Vision backbone.}
We apply the same DINOv3 backbone to $\rgb_t$, $\depth_t$, and $\proxmask_t$ in a batched forward pass. Depth and mask inputs are replicated to three channels and normalized using the same preprocessing as
RGB (after scaling depth to metric units). The DINO feature map is pooled to a fixed-dimensional vector
and projected to the vision embedding used by the world model. As shown in Fig. \ref{fig:dino_vis}, the cosine similarity maps show that the encoder is able to extract spatially meaningful features from the inputs.

\section{Contact-Aware Action Sampling Details}
\label{sec:app-contact-sampling}

\subsection{Active contact set and contact geometry}
\label{app:contact_set_geometry}

We define the active contact set $\cset_t$ from taxels whose measured normal force exceeds threshold $f_{\min}$. For each active taxel $k$, we treat the contact point as the taxel center obtained from forward
kinematics. The corresponding translational Jacobian $\mathbf{J}_{k,t}\in\R^{3\times \ndof}$ is computed
from the robot model. Each taxel has a known outward surface normal in link frame. We transform it to the world frame and negate it to obtain $\nrm_{k,t}$, so the normal points away from the contacted object and positive normal velocity $\Jn_{k,t}\act > 0$ corresponds to separation. This orientation is used by the safety projection (the shaping operators are invariant to row sign) and by the compression update in Appendix~E.

\subsection{Force weighting and projection operators}
\label{app:force_weight_proj}

We use the taxel force magnitude $\fmag_{k,t}$ (scalar normal force) and define
$\contactw_{k,t}=\mathrm{sat}(\fmag_{k,t}/\Fsat)$ to reduce the influence of weak/noisy contacts while
preventing a single contact from dominating via saturation. We compute $\Pforce_t$ using the small matrix inverse
$(\widetilde{\Jn}_t\widetilde{\Jn}_t^\top+\ridge I)^{-1}$ with a stable factorization (e.g., Cholesky).
When $K_t=0$, we set $\Pforce_t=\mathbf{0}$ and $\Pnull_t=I$. The ridge term $\ridge>0$ prevents numerical issues when contact normals are redundant (e.g., nearly collinear rows in $\widetilde{\Jn}_t$).

\begin{table}[h]
\centering
\small
\setlength{\tabcolsep}{6pt}
\renewcommand{\arraystretch}{1.1}
\begin{tabular}{@{}l r@{}}
\textbf{Parameter} & \textbf{Value} \\ \hline
$\sigma_{\mathrm{force}}$ & 1.0 \\
$\sigma_{\mathrm{null}}$  & 0.7 \\
$\sigma_{\min}$           & 0.1 \\
$\ridge$                  & $10^{-4}$ \\
\end{tabular}
\caption{Hyperparameter values for contact-conditioned shaping and safety projection.}
\label{tab:contact_sampling_hparams}
\end{table}

\subsection{Safety projection QP}
\label{app:safety_qp}

We form $\cset^{\mathrm{viol}}_t=\{k\in\cset_t \mid \fmag_{k,t}\ge \Fsafe\}$ and stack the corresponding rows into $\mathbf{J}^{\mathrm{viol}}_{t}$. Because $|\cset^{\mathrm{viol}}_t|$ is small, we solve it efficiently with an active-set method. All $2^K$ active sets are enumerated, the corresponding equality-constrained least-squares problems are solved in closed form, and the feasible solution with minimum objective is selected. If $\cset^{\mathrm{viol}}_t=\emptyset$, we skip the projection.

Table \ref{tab:contact_sampling_hparams} shows the parameters obtained in simulation by commanding the robot to regulate a fixed goal end-effector pose while injecting contacts of varying magnitudes and locations along the arm. We selected values that minimized safety violations and improved recovery behavior, using metrics including contact-limit violations, escape (separation) speed under high force, and tracking error.

The projection assumes contact normals are locally constant over the short MPC horizon, which is
reasonable under our planning frequency. Since the constraint is
applied only to contacts above $\Fsafe$ and enforces separation rather than precise force control, errors
in the assumed normals result in conservative behavior rather than unsafe actions.

\begin{table}[h]
    \centering
    \small
    \setlength{\tabcolsep}{6pt}
    \renewcommand{\arraystretch}{1.2}
    \begin{tabular}{lcc}
        \toprule
        \textbf{Metric} & \textbf{Contact-aware} & \textbf{Contact-aware + QP} \\
        \midrule
        Violations & $84.3 \pm 39.8$ & \textbf{50.0 $\pm$ 20.1} \\
        \bottomrule
    \end{tabular}
    \caption{Ablation results for QP-based projection.}
    \label{tab:violations_contact_aware}
\end{table}

We perform multi-contact ablation experiments to observe the effect of QP-based projection. Table \ref{tab:violations_contact_aware} reports the number of safety force violations under contact-aware control, comparing contact-conditioned shaping alone against shaping combined with our QP-based safety projection.

\section{Hybrid Predictive Model and Planning Details}
  \label{sec:hybrid_model}

  Each timestep is encoded into a latent vector $\lat_t\in\R^{d_z}$ with $d_z=\texttt{384}$.
  We form $\lat_t$ by concatenation of modality embeddings from
  RGB, depth, proximity mask, forces, and proprioception. The predictor conditions on a context window $\lat_{t-\histlen+1:t}$ with $\histlen=\texttt{3}$. We implement $\Fl$ as a
  Transformer/Vision-Transformer-style predictor with
  6 blocks, hidden size 384, and 8 attention heads. We use a frame-causal attention mask so that tokens from timestep $t'$ attend only to timesteps
  $\le t'$ within the context window. The predictor outputs $\latHat_{t+1}$, and we obtain $\latHat_{t+1:t+\horizon}$ via iterative rollout.

  Actions are commanded joint velocities $\act_t\in\R^{\ndof}$. We normalize actions using per-dimension statistics (mean/std) and embed them with an MLP action encoder
  $\phi_{\mathrm{act}}:\R^{\ndof}\rightarrow\R^{d_a}$ with $d_a=\texttt{10}$ and architecture
  1D Conv (kernel=1, stride=1).

  When predicting $\latHat_{t+k}$, we condition on the step-aligned action $\act_{t+k-1}$
  by injecting $\phi_{\mathrm{act}}(\act)$ into each Transformer block via AdaLN, i.e., the action embedding
  produces per-block scale/shift parameters.

  Teacher-forced training uses the ground-truth context $\lat_{t-\histlen+1:t}$ to predict
  $\latHat_{t+1:t+\horizon}$. For the multistep rollout loss, we unroll autoregressively for $k$ steps by
  updating the context with predicted latents using a sliding window:
  append $\latHat$ and drop oldest.
  We use $K_{\mathrm{roll}}=\texttt{2}$ and weights $\lambda_k=\texttt{0.25}$.

  We attach lightweight decoder heads $\psi_m$ to $\latHat_{t+i}$ for planning-relevant signals.
  The binary contact decoder $\decForce$ is a 2-layer MLP with hidden dims [64, 32] and ReLU activations that
  predicts per-taxel contact probabilities through a sigmoid output; we supervise it with a binary
  cross-entropy loss, with ground-truth contacts defined by thresholding measured forces at 2\,N.
  The joint-state decoder predicts $(\q)$ (14D with sin/cos representation for $\q$) with an $\ell_2$ loss.
  These decoded signals are used to derive force estimates for scoring (Appendix~E). We train end-to-end (encoders + predictor + decoders) using AdamW with learning rate $5{\times}10^{-4}$ (predictor), batch size $B=\texttt{32}$, and
  100 epochs. Segments are sampled as $(\histlen+\horizon)$-length windows with random offsets.

  \subsection{Two-fidelity rollouts for real-time control}
  \label{app:two_fidelity_details}

  \paragraph{Draft model design and training}
  The draft predictor $\Fl^{\mathrm{draft}}$ shares the same encoders and action embedding
  $\phi_{\mathrm{act}}$ as the full predictor, but uses a lightweight two-layer MLP. We train $\Fl^{\mathrm{draft}}$ purely using teacher forcing loss (imitate the full model).

  \paragraph{Selection and rescoring}
  At each MPC step $t$, we sample $\Nroll$ action sequences $\{\act^{(j)}_{t:t+\horizon-1}\}_{j=1}^{\Nroll}$.
  We evaluate the analytical rollout $\Fr$ for all $\Nroll$ particles. For the learned rollout, we:
  (i) roll out $\Fl^{\mathrm{draft}}$ for all particles to obtain draft predictions and costs
  $\{ \Cost^{\mathrm{draft}}_j \}_{j=1}^{\Nroll}$,
  (ii) select the top-$M$ candidates
  $\mathcal{I}_M=\texttt{TopM}(\{\Cost^{\mathrm{draft}}_j\},M)$ with $M=10$ and $\mathcal{I}_M\subset\{1,\dots,\Nroll\}$,
  and (iii) re-roll out only those candidates with the full predictor $\Fl$ over the full horizon to compute
  $\{ \Cost^{\mathrm{full}}_j \}_{j\in\mathcal{I}_M}$.

  \paragraph{Draft-error safeguard and fallback policy}
  To detect when the draft model is unreliable, we monitor a one-step latent prediction error at runtime:
  \[
  e_t \;=\; \|\latHat^{\mathrm{draft}}_{t+1} - \lat_{t+1}\|_2,
  \]
  where $\lat_{t+1}=\enc(\obscc_{t+1})$ is computed from the next observation.
  If $e_t>\tau_{\mathrm{draft}}$ with $\tau_{\mathrm{draft}}=0.1$, we disable the two-fidelity
  scheme and evaluate the full model for all particles (equivalently, set $M=\Nroll$) for the next
  $T_{\mathrm{full}}=4$ MPC steps.

\section{Rollout Cost Details}
\label{app:cost_details}

\subsection{Kinematic cost $\ckin$}
\label{app:ckin}

We decompose $\ckin(\hat{s}_{t+i})$ into feasibility and task terms:
\begin{equation}
\begin{aligned}
\ckin(\hat{s}_{t+i})
&=
\lambda_{\mathrm{lim}}\,c_{\mathrm{lim}}(\hat{\q}_{t+i})
+
\lambda_{\mathrm{coll}}\,c_{\mathrm{coll}}(\hat{\q}_{t+i})
+
\lambda_{\mathrm{ee}}\,c_{\mathrm{ee}}(\hat{\vct{e}}_{t+i})\\
&\quad+
\lambda_{u}\,c_{u}(\act_{t+i}),
\label{eq:ckin_decomp_app}
\end{aligned}
\end{equation}

with $\lambda_{\mathrm{lim}}=1000.0$, $\lambda_{\mathrm{coll}}=1000.0$, $\lambda_{\mathrm{ee}}=[15.0,10.0]$
(orientation, position), and $\lambda_{u}=1.0$.

\paragraph{Joint limits}
Let $(\q_{\min},\q_{\max})$ be joint bounds and $\Delta_q$ a soft margin. We use
\begin{equation}
c_{\mathrm{lim}}(\hat{\q})
=
\sum_{m=1}^{\ndof}
\Big(
[\hat{q}_m-(q_{\max,m}-\Delta_q)]_+^2
+
[(q_{\min,m}+\Delta_q)-\hat{q}_m]_+^2
\Big),
\label{eq:joint_limit_cost_app}
\end{equation}
with $\Delta_q = 0.05\,(q_{\max,m}-q_{\min,m})/2$.

\paragraph{Self-collision}
We follow STORM's implementation and learn a lightweight collision classifier for our robot model.

\paragraph{Task-space tracking (if specified)}
When a task-space end-effector target pose $\vct{e}_g=(\vct{p}_g,\mathbf{R}_g)$ is defined, we penalize
translation and rotation:
\begin{equation}
c_{\mathrm{ee}}(\hat{\vct{e}})
=
\frac{\|\hat{\vct{p}}-\vct{p}_g\|_2^2}{\sigma_p^2}
+
\frac{\|\mathbf{I} - \mathbf{R}_g^\top \hat{\mathbf{R}}\|_F^2}{\sigma_R^2}.
\label{eq:ee_cost_app}
\end{equation}

\subsection{Active force cost $c_{\mathrm{active}}$}
\label{app:active_force_cost}

Let $\cset_t$ denote the active contact set at time $t$. For each $k\in\cset_t$, we form a spring-based
force estimate along the measured contact normal using the kinematic rollout $\hat{\q}_{t+1:t+\horizon}$.
For taxel $k$, let $\Jn_{k,t}=\nrm_{k,t}^\top \mathbf{J}_{k,t}$ be the normal Jacobian row. Given a
kinematic rollout $\hat{\q}_{t:t+\horizon}$, define $\Delta\q_{t+i}=\hat{\q}_{t+i}-\hat{\q}_{t+i-1}$ and
the predicted normal displacement
\begin{equation}
\Delta x^n_{k,t+i}=\Jn_{k,t+i}\,\Delta\q_{t+i}.
\label{eq:normal_disp}
\end{equation}
We treat motion into contact as compression and accumulate
\begin{equation}
\hat{\delta}_{k,t+i}=\max\!\big(0,\;\hat{\delta}_{k,t+i-1}-\Delta x^n_{k,t+i}\big),
\qquad \hat{\delta}_{k,t}=0,
\label{eq:compression_update}
\end{equation}
then estimate normal force with a linear spring
\begin{equation}
\hat{F}_{k,t+i}=k_{\mathrm{spr}}\,\hat{\delta}_{k,t+i},
\label{eq:spring_force}
\end{equation}
where $k_{\mathrm{spr}}$ is fit offline. For active contacts, we initialize the spring compression using the measured force at time $t$ to reduce
bias in the rollout. For each $k\in\cset_t$ with measured normal force $\fmag_{k,t}$, we set
\begin{equation}
\hat{\delta}_{k,t}=\mathrm{clip}\!\left(\frac{\fmag_{k,t}}{k_{\mathrm{spr}}},\,0,\,\delta_{\max}\right),
\label{eq:delta_init_measured}
\end{equation}
and then propagate $\hat{\delta}_{k,t+i}$ using Eq.~\eqref{eq:compression_update}. The resulting force
estimate is $\hat{F}_{k,t+i}=k_{\mathrm{spr}}\hat{\delta}_{k,t+i}$. We penalize violations of the safety threshold $\Fsafe$ over the active
contacts with
\begin{equation}
c_{\mathrm{active}}(\hat{\q}_{t+i}, \forces_t)
=
\frac{1}{|\cset_t|}
\sum_{k\in\cset_t}
\big[\hat{F}_{k,t+i}-\Fsafe\big]_+^2.
\label{eq:active_force_cost}
\end{equation}

\subsection{Predicted force cost $c_{\mathrm{pred}}$}
\label{app:cforce}

Instead of decoding raw forces directly from the predicted latents (which can be noisy when evaluated
far into the horizon), we derive a stable force signal for scoring from the decoded binary contacts and
joint positions. We first use the binary contact decoder $\decForce$ to estimate
the contact onset time $\hat{\tau}_k$ for each taxel $k$. We record the predicted joint configuration at
onset, $\hat{\q}_{t+\hat{\tau}_k}$, obtained from the joint-state decoder, and treat it as the reference
configuration for subsequent force estimation. We then apply the same spring model (with stiffness
$k_{\mathrm{spr}}$ as above) to convert predicted compression after onset into a per-taxel normal force
estimate $\forcesHat_{t+i}\in\R^{N_s}$.

We penalize predicted forces above the safety threshold $\Fsafe$ with a hinge-squared penalty:
\begin{equation}
c_{\mathrm{pred}}(\forcesHat_{t+i})
=
\frac{1}{N_s}
\sum_{k=1}^{N_s}
\big[\forcesHat_{k,t+i}-\Fsafe\big]_+^2.
\label{eq:force_cost_app}
\end{equation}

The force safety cost $\cforce$ used in the main paper combines the active and predicted terms:
\begin{equation}
\cforce(\hat{s}_{t+i}, \latHat_{t+i}, \forces_t)
=
c_{\mathrm{active}}(\hat{\q}_{t+i}, \forces_t)
+
c_{\mathrm{pred}}(\forcesHat_{t+i}).
\label{eq:cforce_total_app}
\end{equation}

\subsection{Latent goal cost $\cgoal$}
\label{app:cgoal}

We encode the goal observation as $\latGoal=\enc(\obscc_g)$, where $\obscc_g$ is formed analogously to
$\obscc_t$ (Sec.~\ref{sec:contact_centric_repr}). The goal cost is
\begin{equation}
\cgoal(\latHat_{t+i},\latGoal)=\|\latHat_{t+i}-\latGoal\|_2^2.
\label{eq:latent_goal_cost_app}
\end{equation}

\subsection{Weights and implementation}
\label{app:cost_weights_impl}

Costs are summed over the horizon with discount factor $\gamma=0.98$.
Infeasible rollouts (large joint-limit violations or self-collision) are handled by increasing the
corresponding penalty by a factor of $1000$ rather than rejecting the sample.

\section{IQL Value Learning for Long-Horizon Tasks}
\label{sec:iql_value}

\subsection{Progress Variables and Subgoals}
For each long-horizon task, the dataset provides low-dimensional progress variables
$\xi_t\in\mathbb{R}^{d_g}$ (e.g., body turn angle, limb joint angles). We obtain the values for these indicators during data collection by mounting ArUco markers on the head and limb joints. Note that these markers are not required at test time. Instead, we train progress heads that predict these progress variables for the current $\lat_t$. At test time we specify an ordered
subgoal sequence $\{\xi_g^{(1)},\ldots,\xi_g^{(G)}\}$ and maintain an active index $g$ during execution.
We advance to the next subgoal when the measured progress satisfies
\begin{equation}
\|\xi_t - \xi_g^{(g)}\|_2 \le \delta_\xi,
\label{eq:subgoal_switch}
\end{equation}
for a fixed threshold $\delta_\xi$ (task-dependent).

\subsection{Latent Encoding and Goal Representation}
We freeze the multimodal encoders (Sec.~\ref{sec:contact_centric_repr}) and compute a latent embedding
$\lat_t$ for each observation. For each subgoal $\xi_g^{(g)}$, we also form a corresponding goal
observation and embed it with the same frozen encoders to obtain the goal latent $\latGoal$. The choice of subgoals depends on task difficulty. For instance, in case of rollover, the subgoals are chosen as the two intermediate frames where contact switching happens (robot pulls from near the knee and then moves closer to the lower back and pulls from there).

\subsection{Offline Transition Dataset}
From the expert dataset, we construct an offline transition set
\begin{equation}
\mathcal{D}=\{(\lat_t,\act_t,\lat_{t+1},\latGoal,r_t,d_t)\},
\end{equation}
where $\act_t$ is the executed action, $d_t\in\{0,1\}$ indicates episode termination, and the reward is
defined as the negative of the cost used for scoring task state (Sec.~\ref{sec:hybrid_cost}):
\begin{equation}
r_t \;=\; -\ell_t.
\label{eq:iql_reward}
\end{equation}
$\ell_t$ includes a progress/subgoal tracking term built from the labeled
$\xi_t$ and current subgoal $\xi_g^{(g)}$, e.g.,
\begin{equation}
\ell_{\mathrm{prog}}(\xi_t,\xi_g^{(g)}) = \|\xi_t-\xi_g^{(g)}\|_2^2.
\end{equation}
For rollover, this is the L2 error between the goal head angle and current head angle (predicted by the progress head). In case of limb repositioning, this is the L2 error between goal limb configuration and current limb configuration.

\subsection{Goal-Conditioned IQL Critic}
We implement goal-conditioning by concatenating the state and goal latents (and optionally their difference):
\begin{equation}
x_t = [\lat_t,\;\latGoal,\;(\latGoal-\lat_t)].
\label{eq:iql_features}
\end{equation}
We learn double action-value functions $Q_{\phi_1}(x_t,\act_t)$, $Q_{\phi_2}(x_t,\act_t)$ and a value
function $V_{\theta}(x_t)$, with a slowly-updated target value network $V_{\bar{\theta}}$.

\paragraph{Q-learning objective}
The TD target is
\begin{equation}
y_t = r_t + \gamma(1-d_t)\,V_{\bar{\theta}}(x_{t+1}),
\end{equation}
and we minimize
\begin{equation}
\mathcal{L}_Q =
\mathbb{E}_{\mathcal{D}}\!\left[
\left(Q_{\phi_1}(x_t,\act_t)-y_t\right)^2
+
\left(Q_{\phi_2}(x_t,\act_t)-y_t\right)^2
\right].
\end{equation}

\paragraph{Expectile value objective}
Following IQL, $V_\theta$ is trained via expectile regression toward the in-dataset action values:
\begin{equation}
\mathcal{L}_V =
\mathbb{E}_{\mathcal{D}}\!\left[
\rho_{\tau}\!\left(\min(Q_{\phi_1},Q_{\phi_2})(x_t,\act_t)-V_\theta(x_t)\right)
\right],
\end{equation}
where $\rho_{\tau}$ is the expectile loss with expectile parameter $\tau\in(0,1)$.

We use discount $\gamma=0.99$, expectile $\tau=0.7$, and Polyak target update rate $\alpha=0.005$. All networks ($Q_{\phi_1}, Q_{\phi_2}, V_\theta$, and the AWR actor $\pi_\psi$) are 2-layer MLPs, dropout $p=0.1$, and ReLU activations. We train all networks with AdamW ($\mathrm{lr}=3{\times}10^{-4}$). The advantage-weighted regression temperature is $\beta_{\mathrm{AWR}}=3.0$.

\subsection{Using the Value for Terminal Shaping in MPPI}
At test time, MPPI rolls out the learned latent predictor $\Fl$ to obtain the horizon-end latent
$\latHat_{t+\horizon}$ for each sampled action sequence. We then augment the rollout cost as
\begin{equation}
\Cost_{\mathrm{aug}}(\act_{t:t+\horizon-1})
=
\Cost(\act_{t:t+\horizon-1})
-
\wVal\,\Vfn(\latHat_{t+\horizon}, \latGoal),
\end{equation}
where $\Vfn$ is the learned IQL value function. Since $\Vfn$ estimates reward-to-go and MPPI minimizes cost, the negative sign converts reward-to-go into a cost-to-go term. We set $\wVal=0.1$.

\begin{table}[h]
    \centering
    \small
    \setlength{\tabcolsep}{8pt}
    \renewcommand{\arraystretch}{1.2}
    \begin{tabular}{lc}
        \toprule
        \textbf{Method} & \textbf{SR} \\
        \midrule
        TACTIC & $6/10$ \\
        TACTIC (w/o Value Fn.) & $2/10$ \\
        \bottomrule
    \end{tabular}
    \caption{Ablation on the effect of the terminal value function, reported as success rate (SR) over 10 trials.}
    \label{tab:ablation_value_fn_sr}
\end{table}

\textbf{Ablation with value function.} We ablate the terminal value function in TACTIC to assess its contribution to task success in maze environment, reporting success rate (SR) over 10 trials.

\section{Tactile Skin: Hardware, Frames, Pose/Normal Calibration, and Limitations}
\label{sec:app-tactile}

\subsection{Hardware overview}
We mount 22 Tekscan FlexiForce A502 force-sensing resistors (FSRs) on the robot to measure distributed
normal contact forces along the arm. Each pad measures normal force over a $50.8\,\mathrm{mm}\times
50.8\,\mathrm{mm}$ sensing area and outputs a resistance change proportional to applied load. Sensors are
grouped into four physical mounts: the two wrist joints (5 pads each), the forearm link (6 pads), and
the upper-arm link (6 pads). Each mount is instrumented with an Adafruit Feather M0 WiFi board that
streams raw sensor readings over UDP at 800\,Hz (per board).

\subsection{Sensor readings and preprocessing}
\textbf{Taring.} Before interaction, we tare each sensor by estimating its no-contact offset from a
recent window of raw readings. A tare is accepted only when the window variance is below a stability
threshold.

\textbf{Filtering.} We apply a median filter followed by a low-pass filter to the tared signal.
The resulting tactile vector $\forces_t\in\R^{N_s}$ is published at 100\,Hz for control and logging.

\subsection{Taxel pose, normals, and contact Jacobians}
\textbf{Taxel extrinsics.} For each taxel $k$, we define a taxel frame $\{T_k\}$ attached to the sensing
pad and store its fixed transform with respect to the corresponding link frame $\{L\}$,
${}^{L}\mathbf{T}_{T_k}$. We obtain taxel positions and outward normals in the link frame from manual
measurements (vernier caliper) and CAD alignment, and include these extrinsics in the robot model.

\textbf{Runtime calculation.} At runtime, forward kinematics provides the link pose
${}^{W}\mathbf{T}_L(\q_t)$, which we use to transform each taxel pose and normal into the world frame and
to compute the translational Jacobian at the taxel center. We validate these transforms by visualizing taxel frames across diverse arm configurations.

\subsection{Force calibration}
We calibrate the skin sensor using an ATI Nano25 force/torque sensor. We repeatedly apply normal loads over a
range of magnitudes and fit a linear model that maps raw readings to force in Newtons using
least-squares regression.

\subsection{Driver and data integrity checks}
We implement a \texttt{skin\_driver} that receives UDP packets from the WiFi boards and exposes ROS
services and topics for taring and reading tactile values. The driver includes a watchdog for common
failure modes (e.g., missing packets, low battery, disconnected boards, inactive channels). The driver
performs the filtering discussed above and publishes the filtered forces as an array.

\subsection{Limitations}
\textbf{Coverage.} The tactile skin provides discrete coverage over a subset of the arm surface; contact
can occur in uninstrumented regions. \textbf{Normal-only sensing.} The sensors measure normal force only. This ignores tangential forces and
frictional effects (e.g., shear during sliding), which can be informative for both scoring and sampling
in contact-rich tasks. \textbf{System dependence.} Our method benefits from improved tactile coverage, higher bandwidth, and better calibration. The core algorithm does not assume a specific sensor type, but performance will improve depending on the quality of contact localization and force estimation available at runtime.

\section{Data Collection Details}
\label{sec:app-datacollection}

\subsection{Exoskeleton teleoperation interface}
We collect human demonstration data using a custom exoskeleton teleoperation interface that enables
mirrored whole-arm control of a Kinova Gen3 (7-DoF). Mirroring human arm motion is important for
capturing whole-arm contact strategies relevant to caregiving, including distributing contact across
links and coordinating intermediate-link and end-effector motion. The exoskeleton is a scaled replica
of the robot and matches its kinematic structure using seven Dynamixel XL330-M288-T actuators. An
additional Dynamixel XL330-M077-T actuator in a trigger handle is used to start and stop data
collection. We include safety checks and a paired initialization procedure to ensure consistent
alignment between the exoskeleton and robot before teleoperation.

\subsection{Offline dataset summary}
\label{sec:app-dataset}
During data acquisition, RGB-D, tactile, and proprioceptive observations are recorded synchronously at
40\,Hz. RGB and depth images are captured with an Intel RealSense D455 at 60\,Hz and time-synchronized
to the 40\,Hz logging stream. In the 3D Maze environment, we collected 125 expert and 125 random-play
episodes (each $\sim$30\,s). For Side Rollover, we collected 100 expert and 100 random-play episodes.
For Limb Repositioning, we collected 100 expert and 100 random-play episodes. Episodes in the bed-based
environments have an average duration of $\sim$45\,s. Expert trajectories correspond to successful task
completion (e.g., reaching the specified goals in the maze), while random-play trajectories consist of
exploratory motions that still progress toward task-relevant goal states.

\section{Experiment Details}
\label{sec:app-experiment-details}
\subsection{V-JEPA2 and DINO-WM experiment details}
\noindent \textbf{Finetuning V-JEPA2 and Hybrid Model Experiment.} We load the pre-trained encoders and perform action-conditioned finetuning of the predictor using the tabletop reach dataset. To make the model hybrid, we do FK to compute end-effector samples and add a cost for end-effector goal reaching. This cost forces the planner to blindly move along the line connecting the current state and goal state, thus not reasoning about collision avoidance at all. We combine this with the latent goal reaching cost which intuitively encodes the avoidance behavior seen in the dataset. Upon tuning the weights for 5 different start-goal configurations (not included in testing) we fix the cost weights and run evaluations with models that are trained on different percentages of finetuning data. We also run the experiment with different obstacle shapes and sizes, and obtain similar performance improvements since the kinematics-based policy is independent of these effects in the data.

\end{document}